\documentclass{article}
\PassOptionsToPackage{numbers, compress}{natbib}

\usepackage[preprint]{neurips_2026}

\usepackage[utf8]{inputenc} 
\usepackage[T1]{fontenc}    
\usepackage{hyperref}       
\usepackage{url}            
\usepackage{booktabs}       
\usepackage{amsfonts}       
\usepackage{nicefrac}       
\usepackage{microtype}      
\usepackage{xcolor}         
\usepackage{amsmath, amssymb}
\usepackage{algorithm}
\usepackage{algpseudocode}
\usepackage{graphicx} 
\usepackage{wrapfig}
\usepackage{multirow}
\usepackage[table]{xcolor} 
\usepackage{makecell}
\usepackage{tcolorbox}
\usepackage{xcolor}
\usepackage{graphicx}
\usepackage{enumitem}
\usepackage[rightcaption]{sidecap}

\definecolor{casebg}{RGB}{255,248,230}   
\definecolor{titlebg}{RGB}{217,115,13}    
\definecolor{gtred}{RGB}{220,0,0}         

\tcbuselibrary{skins}
\tcbset{
    casebox/.style={
        colback=casebg,
        colframe=gray!50,
        coltitle=titlebg,
        title={\bfseries Case #1: Qwen2.5-VL 7B},
        fonttitle=\bfseries\color{black},
        boxrule=0.5pt,
        arc=3pt,
        boxed title style={colback=titlebg,arc=3pt,boxrule=0pt},
        left=8pt,right=8pt,top=3pt,bottom=3pt, 
        before upper=\raggedright,
        nobeforeafter, 
    }
}
\usepackage{pifont}
\usepackage{tikz}

\title{Breaking Modality Heterogeneity in Low-Bit Quantization for Large Vision-Language Models}

\author{%
  Yi Zhong$^{1}$,
  \quad
  Haotong Qin$^{2}$,
  \quad
  Xindong Zhang$^{3}$,
  \quad
  Lei Zhang$^{3,4}$,
  \quad
  Guolei Sun$^{1}$\thanks{Corresponding author: Guolei Sun, sunguolei.kaust@gmail.com.}
  \\[3pt]
  $^1$ VCIP, College of Computer Science, Nankai University \\ \quad $^2$ D-ITET, ETH Zürich \quad  $^3$ OPPO Research Institute \\
  \quad $^4$ Department of Computing, Hong Kong Polytechnic University\\
  \\[-8pt]
}

\begin{document}

\maketitle

\begin{abstract}
Low-bit post-training quantization (PTQ) is a pivotal technique for deploying Vision-Language Models (VLMs) on resource-constrained devices. However, existing PTQ methods often degrade VLMs' accuracy due to the heterogeneous activation distributions of text and vision modalities during quantization. 
We find that this cross-modal heterogeneity is distributed unevenly across channels: a small subset of channels contains most modality-specific outliers, and these outliers typically reside in different channels for each modality. Motivated by this, we propose \textbf{SplitQ}, a channel-\textbf{Split}ting-driven post-training \textbf{Q}uantization framework. At its core, SplitQ introduces a novel \textit{Modality-specific Outlier Channel Decoupling} (MOCD) module that effectively isolates salient modality-specific outlier channels with minimal overhead. To further address the remaining cross-modal distribution discrepancies, we design an \textit{Adaptive Cross-Modal Calibration} (ACC) module that employs dual lightweight learnable branches to dynamically mitigate modality-induced quantization errors.
Extensive experiments on popular VLMs demonstrate that SplitQ significantly outperforms existing approaches across 6 popular multi-modal datasets under all evaluated quantization settings, including W4A8, W4A4, W3A3, and W3A2. Notably, SplitQ preserves 93.5\% of FP16 performance under the challenging W3A3 setting (69.5 vs. 74.3), pushing the efficiency frontier for deploying advanced VLMs. Our code is available at \url{https://github.com/EMVision-NK/SplitQ}.
\end{abstract}

\section{Introduction}
Vision-Language Models (VLMs)~\cite{bai2023qwen,bai2025qwen25vltechnicalreport,liu2024improved,liu2023visual} have emerged as foundation models for cross-modal understanding, enabling diverse applications such as visual question answering~\cite{antol2015vqa,li2016visual}, image captioning~\cite{you2016image,herdade2019image}, and multimodal reasoning~\cite{lu2022learn,yang2025r1}. However, the massive scale and high computational demands of VLMs pose significant challenges for deployment on resource-constrained devices. Post-Training Quantization (PTQ)~\cite{li2025mbq,hu2026masquant,liu2021post,shang2023post,yao2022zeroquant} has become a popular approach to reduce memory footprints and computational FLOPs. In particular, transformation-based PTQ methods adopt scaling or orthogonal transformations to redistribute weight and activation values, thereby improving quantization accuracy~\cite{ashkboos2024quarot,sun2024flatquant,liu2024spinquant,xiao2023smoothquant,ma2024affinequant}.

While transformation-based PTQ methods are well-established for pure Large Language Model (LLM) quantization, extending them to VLMs remains challenging due to the heterogeneous activation distributions across text and vision modalities. Our empirical analysis reveals that this cross-modal heterogeneity is uneven across channels and exhibits a clear channel-wise structural distribution. 
Specifically, text and vision activations exhibit severe modality-specific outliers in a small number of channels. What's more, the outlier channels of the two modalities are different. 
Existing approaches~\cite{li2025mbq} typically learn a single transformation shared by both modalities across all channels, where the cross-modal heterogeneity can severely distort the optimization objective. Some methods~\cite{yu2025mquant,hu2026masquant} also attempt to handle the two modalities separately, but such treatment does not fully consider the modality-specific channel outliers.

Motivated by our observations, we propose \textbf{SplitQ}, a novel modality-aware framework for VLM quantization. At its core, SplitQ introduces a novel \textit{Modality-specific Outlier Channel Decoupling} (MOCD), which analyzes unique activation patterns (magnitude saliency for vision tokens and response consistency for text tokens) to explicitly divide channels into three disjoint sets: modality-compatible main channels, text-specific outlier channels, and vision-specific outlier channels. The activation and weight matrices are subsequently split according to these channel sets and processed through different quantization paths. 
Although the modality-compatible main channels exhibit significantly reduced heterogeneity, residual cross-modal distribution discrepancies persist.
To address this, SplitQ is further equipped with an \textit{Adaptive Cross-Modal Calibration} (ACC) module which introduces two lightweight learnable branches to mitigate the residual cross-modal quantization errors: one compensates for weight-side errors amplified by cross-modal shifts, while the other specifically recovers activation-side deviations for text tokens.

Extensive experiments on mainstream VLMs demonstrate that SplitQ consistently achieves state-of-the-art performance across various quantization configurations (W4A8, W4A4, W3A3 and W3A2), significantly outperforming existing approaches. Our main contributions are summarized as follows:
\begin{itemize}
\vspace{-0.22cm}
\item We investigate modality heterogeneity in VLM quantization and reveal that cross-modal distribution shifts are highly channel-dependent. Motivated by this, we propose SplitQ with a novel MOCD module to explicitly separate channels into three groups: modality-compatible main channels, text-specific channels, and vision-specific channels.
\vspace{-0.1cm}
\item To further reduce cross-modal quantization errors in the main channels, we equip SplitQ with an ACC module which exploits two lightweight learnable branches for weight smoothing and activation compensation.
\vspace{-0.1cm}
\item We achieve SOTA quantization results on six benchmarks across Qwen2.5-VL-7B and Qwen2.5-VL-3B. SplitQ nearly preserves FP16 performance at W4A4, achieving average scores of 73.5 and 69.6 on the two models, compared with 74.3 and 70.0 for FP16, respectively, and remains effective even at W3A3 and W3A2.
\end{itemize}

\section{Related Work}
\label{gen_inst}
\subsection{LLM Quantization}
Current LLM quantization methods can be broadly categorized according to the quantization stage into quantization-aware training (QAT)~\cite{bondarenko2024low, chen2025efficientqat, liu2024llm,nagel2022overcoming} and post-training quantization (PTQ)~\cite{song2025achieving,wang2026sliderquant,sharify2024post,yao2022zeroquant,ding2023cbq}. Although QAT usually achieves better accuracy, it requires additional training and substantial computational resources, making PTQ a more practical and widely studied paradigm for efficient LLM deployment. Existing PTQ methods can be further divided into weight-only quantization~\cite{lin2024awq, frantar2022gptq, zheng2024mixllm, hu2025moequant, li2024norm, chee2023quip, xu2025rwkvquant, dettmers2023spqr, kim2023squeezellm} and weight-activation quantization~\cite{ma2024affinequant, zhao2025aser, sun2024flatquant, li2023fptq, hu2024llm, shao2023omniquant, hu2025ostquant, wei2022outlier, zhang2024qqq, ashkboos2024quarot, xiao2023smoothquant, liu2024spinquant, yao2023zeroquant}. Weight-only quantization reduces the precision of model weights while keeping activations in high precision, thereby significantly reducing memory. In contrast, weight-activation quantization quantizes both weights and activations, which can reduce memory consumption and improve inference efficiency. Among weight-only methods, GPTQ~\cite{frantar2022gptq} leverages approximate second-order information to perform accurate one-shot weight quantization, while AWQ~\cite{lin2024awq} identifies salient weight channels by observing activation distributions and searches for optimal per-channel scaling to protect them during quantization. For weight-activation quantization, SmoothQuant~\cite{xiao2023smoothquant} smooths activation outliers by migrating quantization difficulty from activations to weights, enabling effective weight-activation quantization. OmniQuant~\cite{shao2023omniquant} introduces learnable weight clipping and equivalent transformation to handle extreme values of weights and activation outliers, and SpinQuant~\cite{liu2024spinquant} learns rotation matrices to reduce outliers in weights and activations.

\begin{figure}[tb]
  \centering
   \includegraphics[width=0.99\linewidth]{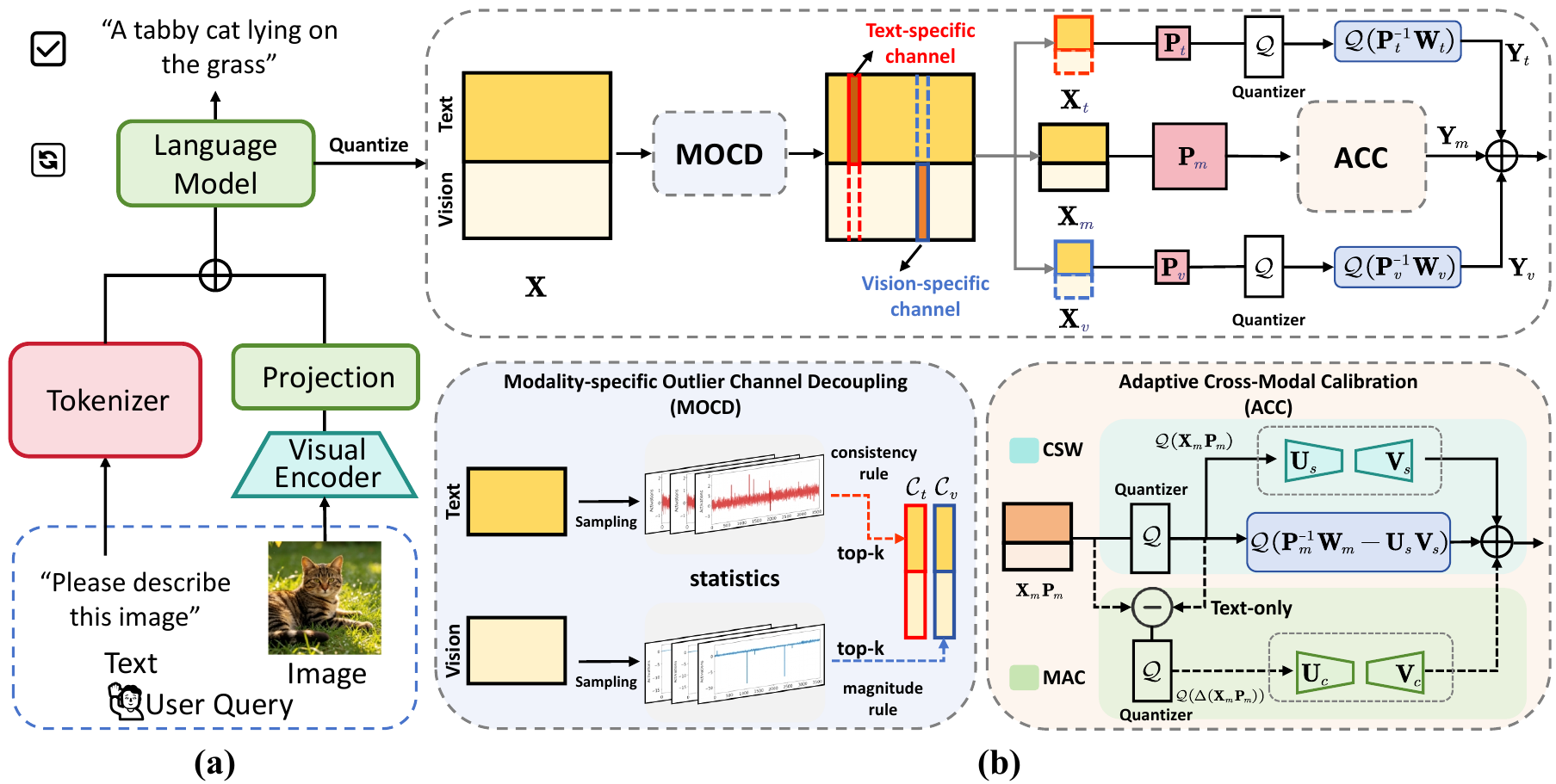}
   \caption{(a) VLM inference pipeline. (b) An overview of SplitQ framework. The two key components of SplitQ include modality-specific outlier channel decoupling (MOCD) module to separate modality-specific outlier channels ($\mathcal{C}_t$ and $\mathcal{C}_v$), and adaptive cross-modal calibration module to further reduce quantization errors caused by cross-modal distribution discrepancies.
   }
   \vspace{-5pt}
   \label{fig:re_pic1}
\end{figure}

\subsection{VLM Quantization}
As quantization techniques for LLM have become increasingly mature, recent studies have started to extend quantization techniques to VLM, where multimodal inputs introduce additional challenges such as modality-dependent activation distributions, redundant visual tokens, and large attention-cache overhead. To address these challenges, several VLM-oriented quantization methods have been proposed. QSLAW~\cite{xie2024advancing} uses quantization-aware scale learning to mitigate quantization errors caused by multimodal activation outliers; Q-VLM~\cite{wang2024q} uses activation entropy as a proxy for block partitioning to balance discretization error and search cost under cross-layer error dependency. MBQ~\cite{li2025mbq} incorporates the different sensitivities across modalities during the calibration process; MQuant~\cite{yu2025mquant} assigns modality-specific static scaling factors to visual and textual tokens; VLMQ~\cite{xue2025vlmq} uses gradient-driven token importance to preserve salient tokens and suppress redundant visual tokens; MASQuant~\cite{hu2026masquant} learns modality-aware smoothing factors to alleviate smoothing misalignment. QSVD~\cite{wang2025qsvd} applies SVD to the combined weight matrices of the query, key, and value, reducing KV-cache size and attention computation overhead; WSVD~\cite{wang2026wsvd} performs each head SVD with local weighted finetuning to achieve practical decoding latency reduction while preserving accuracy. Bi-VLM~\cite{wang2026bi} further explores saliency-aware hybrid quantization for ultra-low-bit precision.

\section{Transformation-based PTQ}
PTQ reduces the computational and memory footprint of models by mapping high-precision floating-point tensor $\mathbf{x}$ to low-precision $N$-bit integer tensor $\hat{\mathbf{x}}_N$:
{\small
\begin{equation}
\hat{\mathbf{x}}_N = Q(\mathbf{x}) = \left( \text{clamp}\left( \left\lfloor \frac{\mathbf{x}}{S_x} \right\rceil + z, q_{\min}, q_{\max} \right) - z \right) \cdot S_x,
\label{eq:quant_def}
\end{equation}
}
where $S_x$ and $z$ denote the scale factor and zero-point, respectively. $\lfloor \cdot \rceil$ is the rounding-to-nearest operator, and $\text{clamp}$ clips values outside the range $[q_{\min}, q_{\max}]$. We use $\text{W}x\text{A}y$ to represent the weight-activation quantization configuration of $x$-bit weights and $y$-bit activations (e.g., W4A4). 

Due to severe outliers in LLM or VLM activations, recent approaches~\cite{sun2024flatquant,liu2024spinquant,xiao2023smoothquant,ashkboos2024quarot} introduce Transformation-based PTQ (TPTQ), where a learnable transformation matrix (e.g., diagonal or orthogonal) is adopted to transfer these outliers from activations to weights, thereby improving quantization friendliness. Specifically, for a linear layer $\mathbf{Y} = \mathbf{XW}$, where $\mathbf{X} \in \mathbb{R}^{T \times D_{\text{in}}}$ and $\mathbf{W} \in \mathbb{R}^{D_{\text{in}} \times D_{\text{out}}}$, the layer can be reformulated based on computational invariance as:
{\small
\begin{equation}
\mathbf{Y} = (\mathbf{X}\mathbf{P}) \cdot (\mathbf{P}^{-1}\mathbf{W}),
\label{eq:comp_invariance}
\end{equation}
}
where $\mathbf{P}$ is a linear optimization matrix that reshapes feature distributions to facilitate adaptive quantization. To optimize learnable parameters,  the objective is defined as the reconstruction loss:
{\small
\begin{equation}
L = \mathcal{L}\left( Q(\mathbf{X}\mathbf{P}) \cdot Q(\mathbf{P}^{-1}\mathbf{W}), \mathbf{XW} \right).
\label{eq:recon_loss}
\end{equation}
}

\begin{algorithm}[t]
\small
\caption{SplitQ linear layer.}
\label{alg:splitq_linear}
\begin{algorithmic}[1]
\Require Activation $\mathbf{X}$, weight $\mathbf{W}$, quantizer $\mathcal{Q}(\cdot)$, 
channel sets $\mathcal{C}_{m}, \mathcal{C}_{t}, \mathcal{C}_{v}$, 
transforms $\mathbf{P}_{m}, \mathbf{P}_{t}, \mathbf{P}_{v}$, 
learnable low-rank branches $(\mathbf{U}_{s}, \mathbf{V}_{s})$ and $(\mathbf{U}_{c}, \mathbf{V}_{c})$.
\Ensure Quantized output $\mathbf{Y}$.
\State Split $\mathbf{X}$ and $\mathbf{W}$ by $\mathcal{C}_{m}, \mathcal{C}_{t}, \mathcal{C}_{v}$:
$(\mathbf{X}_{m}, \mathbf{X}_{t}, \mathbf{X}_{v}),\quad
(\mathbf{W}_{m}, \mathbf{W}_{t}, \mathbf{W}_{v}).$
\State Compute text-specific path:
$\mathbf{Y}_{t}
=
\mathcal{Q}(\mathbf{X}_{t}\mathbf{P}_{t})
\mathcal{Q}(\mathbf{P}_{t}^{-1}\mathbf{W}_{t}).$
\State Compute vision-specific path:
$\mathbf{Y}_{v}
=
\mathcal{Q}(\mathbf{X}_{v}\mathbf{P}_{v})
\mathcal{Q}(\mathbf{P}_{v}^{-1}\mathbf{W}_{v}).$
\State Quantize the main activation:
$\widehat{\mathbf{X}}_{m}=\mathcal{Q}(\mathbf{X}_{m}\mathbf{P}_{m}).$
\If{Cross-modal Weight Smoothing is enabled}
    \State Compute main path:
    $\mathbf{Y}_{m}
    =
    \widehat{\mathbf{X}}_{m}
    \mathcal{Q}(\mathbf{P}_{m}^{-1}\mathbf{W}_{m}-\mathbf{U}_{s}\mathbf{V}_{s})
    +
    \widehat{\mathbf{X}}_{m}
    \mathcal{Q}(\mathbf{U}_{s})
    \mathcal{Q}(\mathbf{V}_{s}).$
\Else
    \State Compute main path:
    $\mathbf{Y}_{m}
    =
    \widehat{\mathbf{X}}_{m}
    \mathcal{Q}(\mathbf{P}_{m}^{-1}\mathbf{W}_{m}).$
\EndIf
\If{Modality-specific Activation Compensation is enabled}
    \State Compensate text tokens:
    $\mathbf{Y}_{m}^{\text{text}}
    \leftarrow
    \mathbf{Y}_{m}^{\text{text}}
    +
    \mathcal{Q}\left(\mathbf{X}_{m}\mathbf{P}_{m}-\widehat{\mathbf{X}}_{m}\right)^{\text{text}}
    \mathcal{Q}(\mathbf{U}_{c})
    \mathcal{Q}(\mathbf{V}_{c}).$
\EndIf
\State Merge three paths:
$\mathbf{Y}=\mathbf{Y}_{m}+\mathbf{Y}_{t}+\mathbf{Y}_{v}.$
\State \Return $\mathbf{Y}$.
\end{algorithmic}
\vspace{-0pt}
\end{algorithm}

\section{Methodology}
\label{sec:methodology}
In this section, we present our SplitQ framework, a PTQ approach for VLMs. Fig.~\ref{fig:re_pic1} provides an overview of the framework, while the core quantized linear layer is detailed in Alg.~\ref{alg:splitq_linear}. 
We first introduce our motivations in \S\ref{subsec:method_motivation}, followed by modality-specific outlier channel decoupling in \S\ref{subsec:channel_decoupling}. Finally, we explain our adaptive cross-modal calibration module in \S\ref{subsec:adaptive_cross_modal_calibration}.

\subsection{Motivations}
\label{subsec:method_motivation}

\begin{wrapfigure}{r}{0.55\textwidth}
  \centering
  \vspace{-10pt}
  \includegraphics[width=\linewidth]{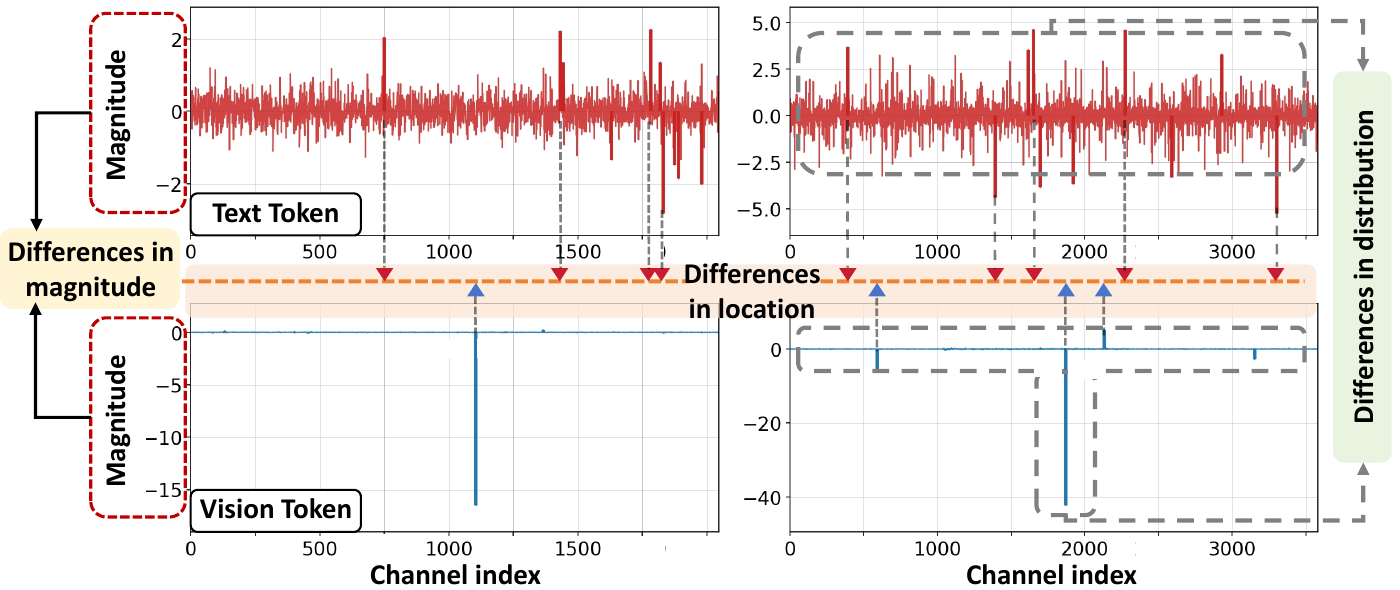}
  \vspace{-15pt}
  \caption{
   Distributions of text and vision activations across channels. \textit{Left}: layer 5 of Qwen2.5-VL 3B; \textit{Right}: layer 10 of Qwen2.5-VL 7B. Notable divergence in distributions underscores the modality heterogeneity.  
  }
  \vspace{-15pt}
  \label{fig:fb}
\end{wrapfigure}

\vspace{-5pt}\paragraph{Modality-specific Outlier Channels.}
We visualize the distributions of text and vision tokens across channels in Fig.~\ref{fig:fb} to investigate the fundamental challenges of quantizing heterogeneous modalities. Our key observations are as follows. 
{\ding{192}} The activation distributions of different modalities exhibit 
fundamental differences. Vision activations usually follow a long-tailed distributions where only a few channels contain extremely large magnitudes, i.e., outliers that need to be handled carefully. In contrast, text activation outliers are distributed more evenly across channels. {\ding{193}} For both modalities, outlier channels account for only a small fraction of all channels. {\ding{194}} More importantly, the activation outliers of text and vision modalities are locate in different channels. This separation provides us an opportunity to disentangle the channel outliers for different modalities.

From observation {\ding{192}}, we can identify a key reason why PTQ for VLM is particularly difficult. In existing transformation-based PTQ methods where the activations of all channels are optimized with a shared transformation matrix $\mathbf{P}$, the modality-specific outlier channels significantly interfere with each other, leading to suboptimal transformations for both modalities.

Inspired by observations {\ding{193}} and {\ding{194}}, we propose a Modality-specific Outlier Channel Decoupling (MOCD) technique to disentangle the interference of modality-specific outliers by separating the text-specific and vision-specific outlier channels out of all channels, resulting in three groups of channels: vision-specific, text-specific and modality-compatible channels. 
Specifically, let $\mathcal{C}=\{0,1,...,D_{\text{in}}\}$ denote all channel indexes. MOCD partitions the input channels into three disjoint sets:
{\small
\begin{equation}
    \mathcal{C}
    =
    \mathcal{C}_{m}
    \cup
    \mathcal{C}_{t}
    \cup
    \mathcal{C}_{v},
    \qquad
    \mathcal{C}_{m}\cap \mathcal{C}_t=\emptyset,
    \qquad
    \mathcal{C}_{m}\cap \mathcal{C}_v=\emptyset,
    \qquad
\end{equation}
}
where $\mathcal{C}_{m}$, $\mathcal{C}_{t}$, and $\mathcal{C}_{v}$ denote modality-compatible, text and vision channels, respectively. 
They also satisfy the following criterion since the number of outlier channels is much smaller than $D_{\text{in}}$:
{\small
\begin{equation}
    |\mathcal{C}_{m}| \gg |\mathcal{C}_{t}|,
    \qquad
    |\mathcal{C}_{m}| \gg |\mathcal{C}_{v}|,
    \label{eq:c-cardi}
\end{equation}
}
where $|\text{set}|$ denotes the cardinality of the set.
Accordingly, the activation/weight matrices are split as:
{\small
\begin{equation}
    \mathbf{X}
    \rightarrow
    \{\mathbf{X}_{m}, \mathbf{X}_{t}, \mathbf{X}_{v}\},
    \qquad
    \mathbf{W}
    \rightarrow
    \{\mathbf{W}_{m}, \mathbf{W}_{t}, \mathbf{W}_{v}\}.
    \label{eq:x-split}
\end{equation}
}

Three matrix pairs $\{\mathbf{X}_{m},\mathbf{W}_{m}\}$, $\{\mathbf{X}_{t},\mathbf{W}_{t}\}$, $\{\mathbf{X}_{v},\mathbf{W}_{v}\}$ are processed differently by learning different transformations. The details of MOCD are further elaborated in \S\ref{subsec:channel_decoupling}.

\vspace{-10pt}\paragraph{Cross-modal Low-bit Quantization Difficulty.}
The heterogeneity between vision and text persists beyond the obvious outlier channels. Even after decoupling these outliers, the remaining channels still exhibit distribution differences across modalities. More specifically, the activation $\mathbf{X}_m$ and weight $\mathbf{W}_m$ still contain mixed distributions of text and vision modalities. 

While high-precision quantization can tolerate such variations, low-bit quantization is highly sensitive to them, as demonstrated in our experiments (\S\ref{sec:experi}). Starting from the baseline FlatQuant~\cite{sun2024flatquant}, we add the proposed MOCD and use an affine transformation matrix to quantize $\{\mathbf{X}_m, \mathbf{W}_m\}$. This approach (FlatQuant+MOCD) gives excellent results at W4A4, but its performance is unsatisfactory at challenging W3A3 and W3A2. To mitigate the subtle distribution differences among modalities within $\mathbf{X}_m$, we design an Adaptive Cross-modal Calibration (ACC) module to further improve the quantization of $\{\mathbf{X}_{m},\mathbf{W}_{m}\}$. Details of the ACC module are provided in \S\ref{subsec:adaptive_cross_modal_calibration}.

\subsection{Modality-specific Outlier Channel Decoupling}
\label{subsec:channel_decoupling}

As discussed above, the input channels $\mathcal{C}$ is split into $\mathcal{C}_m$, $\mathcal{C}_t$, and $\mathcal{C}_v$. In this part, we describe the process of selecting text- and vision- specific outlier channels in detail.

\vspace{-10pt}\paragraph{Modality-aware Channel Selection.}

We first select vision-specific outlier channels by magnitude, followed by identifying text-specific outlier channels via a consistency-based calibration proxy. Let $\mathcal{T}_{v}$ and $\mathcal{T}_{t}$ denote the sampled vision and text tokens, respectively, and let $\mathcal{C}$ be the full channel set. For vision tokens, we score each channel by its maximum absolute activation:
{\small
\begin{equation}
    s_v(c)=\max_{i\in\mathcal{T}_{v}} |X_{i,c}|.
    \label{eq:vision_score}
\end{equation}
}
The vision-specific outlier set is obtained by selecting the top-$K_v$ channels:
{\small
\begin{equation}
    \mathcal{C}_{v}
    =
    \operatorname{TopK}_{c\in\mathcal{C}}
    \left(s_v(c),K_v\right).
    \label{eq:vision_select}
\end{equation}
}
After this, we denote the remaining channels as $\mathcal{C}'=\mathcal{C}\setminus\mathcal{C}_{v}$. For text tokens, absolute magnitude is less indicative, as outlier channels are related to unstable relative responses across tokens. We therefore use the percentile rank within each token as a scale-insensitive measure of channel importance:
{\small
\begin{equation}
    r_{i,c}
    =
    \frac{1}{|\mathcal{C}'|}
    \sum_{j\in\mathcal{C}'}
    \mathbb{I}
    \left(
    |X_{i,j}|\le |X_{i,c}|
    \right),
    \quad
    i\in\mathcal{T}_{t},\ c\in\mathcal{C}' .
    \label{eq:text_percentile}
\end{equation}
}
This rank-based measure suppresses token-wise scale variation and makes channel responses comparable across text tokens. For each channel, we cluster its rank sequence across text tokens into $K$ groups and use the within-cluster variance as a measure of response instability. Let $z_{i,c}$ be the cluster assignment of
$r_{i,c}$ and $\mu_{c,z_{i,c}}$ be the corresponding cluster center:
{\small
\begin{equation}
    s_t(c)
    =
    \frac{1}{|\mathcal{T}_{t}|}
    \sum_{i\in\mathcal{T}_{t}}
    \left(
    r_{i,c}-\mu_{c,z_{i,c}}
    \right)^2 .
    \label{eq:text_score}
\end{equation}
}
A larger $s_t(c)$ suggests less stable relative responses across text tokens. We select the top-$K_t$ channels as text-specific channels:
{\small
\begin{equation}
    \mathcal{C}_{t}
    =
    \operatorname{TopK}_{c\in\mathcal{C}'}
    \left(s_t(c),K_t\right).
    \label{eq:text_select}
\end{equation}
}
The remaining channels form the modality-compatible main channel set:
{\small
\begin{equation}
    \mathcal{C}_{m}
    =
    \mathcal{C}
    \setminus
    \left(
    \mathcal{C}_{v}\cup\mathcal{C}_{t}
    \right).
    \label{eq:main_channel_set}
\end{equation}
}

\vspace{-10pt}\paragraph{Quantization Calibration.} Following channel separation, activation $\mathbf{X}$ and weight $\mathbf{W}$ are split into $\{\mathbf{X}_{m}, \mathbf{X}_{t}, \mathbf{X}_{v}\}$ and $\{\mathbf{W}_{m}, \mathbf{W}_{t}, \mathbf{W}_{v}\}$, respectively. The two matrix groups $\{\mathbf{X}_{t}, \mathbf{W}_{t}\}$ and $\{\mathbf{X}_{v}, \mathbf{W}_{v}\}$ correspond to text- and vision- specific outlier channels. They are processed with independent modality-aware transformations:
{\small
\begin{equation}
    \mathbf{Y}_{t}
    =
    Q(\mathbf{X}_{t}\mathbf{P}_{t})
    Q(\mathbf{P}_{t}^{-1}\mathbf{W}_{t}),
    \label{eq:text_outlier}
\end{equation}
}
{\small
\vspace{-10pt}
\begin{equation}
    \mathbf{Y}_{v}
    =
    Q(\mathbf{X}_{v}\mathbf{P}_{v})
    Q(\mathbf{P}_{v}^{-1}\mathbf{W}_{v}).
    \label{eq:vision_outlier}
\end{equation}
}
Here, $\mathbf{P}_{t}$ and $\mathbf{P}_{v}$ are optimized for the text- and vision-specific outlier channels, respectively. 
The main matrix group $\{\mathbf{X}_{m}, \mathbf{W}_{m}\}$ corresponding to modality-compatible channels is processed by our adaptive cross-modal calibration module in \S\ref{subsec:adaptive_cross_modal_calibration}.

\subsection{Adaptive Cross-modal Calibration}
\label{subsec:adaptive_cross_modal_calibration}
As detailed above, MOCD isolates modality-specific outliers, which are then processed through independent paths (Eq.~\ref{eq:text_outlier}, \ref{eq:vision_outlier}). The remaining channels constitute the main activation-weight pair
\begin{wrapfigure}{r}{0.5\textwidth}
  \centering
  \vspace{-5pt}
  \includegraphics[width=\linewidth]{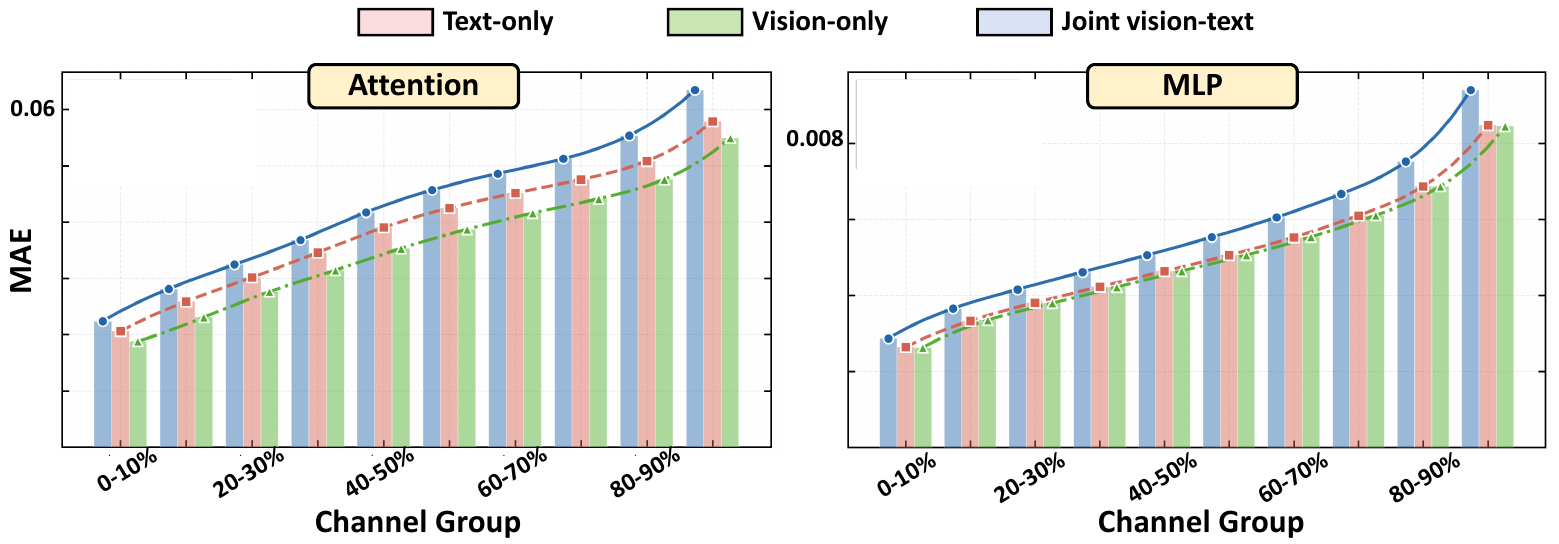}
  \vspace{-10pt}
  \caption{
  {Comparison of weight quantization error $\Delta(\mathbf{P}_{m}^{-1}\mathbf{W}_{m})$ across channels using activations of different modalities: text-only, vision-only, and joint vision-text. The $x$-axis (Channel Group) partitions all channels, sorted in ascending order of quantization error MAE, into 10 equal-sized bins (each containing 10\% of channels). The results show that vision-text activations consistently amplify this error term.}
  }
  \label{fig:pw}
  \vspace{-5pt}
\end{wrapfigure}
$\{\mathbf{X}_m, \mathbf{W}_m\}$. We can adopt a shared transformation matrix $\mathbf{P}_m$ for quantization as follows:
{\small
\begin{equation}
    \mathbf{Y}_{m}
    =
    Q(\mathbf{X}_{m}\mathbf{P}_{m})
    Q(\mathbf{P}_{m}^{-1}\mathbf{W}_{m}).
    \label{eq:main-path}
\end{equation}
}

$\{\mathbf{X}_m, \mathbf{W}_m\}$ encompass the majority channels of original $\{\mathbf{X}, \mathbf{W}\}$. Although these channels are relatively compatible after outlier decoupling, their underlying cross-modal distribution differences persist. Forcing a unified transformation $\mathbf{P}_m$ over such a large-scale channel space inevitably leads to sub-optimal smoothing for both modalities. Consequently, this amplifies the weight-side quantization error $\Delta(\mathbf{P}_{m}^{-1}\mathbf{W}_{m})$ and activation-side quantization error $\Delta(\mathbf{X}_{m}\mathbf{P}_{m})$.
Here, we denote $\Delta(\mathbf{M})=\mathbf{M}-Q(\mathbf{M})$ to represent the quantization error of $\mathbf{M}$. 
To solve this issue, we introduce the ACC module, achieving fine-grained calibration.

\vspace{-10pt}\paragraph{Cross-modal Weight Smoothing (CWS).}
As shown in Fig.~\ref{fig:pw}, the main path still exhibits multimodal sensitivity. By comparing PTQ results calibrated on single-modal versus multi-modal tokens, we observe that the weight-side error $\Delta(\mathbf{P}_{m}^{-1}\mathbf{W}_{m})$ is significantly amplified by the distributional differences between vision and text activations. To mitigate this cross-modal conflict, instead of quantizing the weight $\mathbf{P}_{m}^{-1}\mathbf{W}_{m}$ directly, we extract a low-rank adaptive component $\mathbf{U}_{s}\mathbf{V}_{s}$ to absorb these sensitive patterns:
{\small
\begin{equation}
    \mathbf{P}_{m}^{-1}\mathbf{W}_{m}
    =
    \left(
    \mathbf{P}_{m}^{-1}\mathbf{W}_{m}
    -
    \mathbf{U}_{s}\mathbf{V}_{s}
    \right)
    +
    \mathbf{U}_{s}\mathbf{V}_{s},
    \label{eq:weight_decomp}
\end{equation}
}
where $\mathbf{U}_{s}\in\mathbb{R}^{D_m \times r}$ and $\mathbf{V}_{s}\in\mathbb{R}^{r \times D_{\text{out}}}$ are learnable low-rank matrices. Let $\widehat{\mathbf{X}}_{m}=Q(\mathbf{X}_{m}\mathbf{P}_{m})$ denote the quantized input. According to Eq.~\ref{eq:weight_decomp}, we explicitly decouple $\mathbf{P}_{m}^{-1}\mathbf{W}_{m}$ into a main residual part $(\mathbf{P}_{m}^{-1}\mathbf{W}_{m} - \mathbf{U}_{s}\mathbf{V}_{s})$ and a low-rank part $\mathbf{U}_{s}\mathbf{V}_{s}$. By quantizing and computing these components independently, the final output of the main path is formulated as:
{\small
\begin{equation}
    \mathbf{Y}_{m}
    =
    \widehat{\mathbf{X}}_{m}
    Q\left(
    \mathbf{P}_{m}^{-1}\mathbf{W}_{m}
    -
    \mathbf{U}_{s}\mathbf{V}_{s}
    \right)
    +
    \widehat{\mathbf{X}}_{m}
    Q(\mathbf{U}_{s})
    Q(\mathbf{V}_{s}).
    \label{eq:weight_smooth}
\end{equation}
}
While \cite{li2024svdquant} adopts a similar low-rank architecture to absorb main energy or outliers in single-modal scenarios, our structural intent is distinct. We utilize the low-rank branch $\mathbf{U}_{s}\mathbf{V}_{s}$ adaptively to absorb the weight variations triggered by cross-modal activation shifts. By extracting these sensitive cross-modal components, the remaining main weight $(\mathbf{P}_{m}^{-1}\mathbf{W}_{m} - \mathbf{U}_{s}\mathbf{V}_{s})$ exhibits a smoother distribution, allowing for low-error quantization.

\vspace{-10pt}\paragraph{Modality-specific Activation Compensation (MAC).}
While CWS reduces the weight-side quantization error caused by cross-modal heterogeneity, the activation-side remains more challenging. Since the shared transformation $\mathbf{P}_m$ is optimized under conflicting modality distributions, its smoothing effect is inevitably compromised, leaving a non-negligible activation residual $\Delta(\mathbf{X}_{m}\mathbf{P}_{m})$. Unlike static weights, activations are input-dependent and dynamically vary across modalities, making it difficult to reformulate their quantization-sensitive components through structural decomposition (as we did in Eq.~\ref{eq:weight_smooth}). We therefore introduce a direct compensation branch to recover the output deviation caused by activation quantization:
{\small
\begin{equation}
    \Delta \mathbf{Y}_{m}^{\text{act}} =
    \Delta(\mathbf{X}_{m}\mathbf{P}_{m})
    \mathbf{P}_{m}^{-1}\mathbf{W}_{m},
    \label{eq:act_error}
\end{equation}
}
where $\Delta \mathbf{Y}_{m}^{\text{act}}$ is the output deviation induced by the activation residual.

As revealed by prior observations~\cite{yu2025mquant}, text activations are more sensitive to quantization noise due to their dense semantic nature, whereas vision activations are relatively redundant. Therefore, text compensation usually captures the more critical activation-side errors. For efficient implementation, we use learnable low-rank matrices $\mathbf{U}_{c}\mathbf{V}_{c}$ to approximate the weight mapping $\mathbf{P}_{m}^{-1}\mathbf{W}_{m}$ in Eq.~\ref{eq:act_error}. Furthermore, all factors in this branch are fully quantized, and we apply this compensation only to the text-side output $\mathbf{Y}_{m}^{\text{text}}$:
{\small
\begin{equation}
    \mathbf{Y}_{m}^{\text{text}}
    \leftarrow
    \mathbf{Y}_{m}^{\text{text}}
    +
    Q\big(\Delta(\mathbf{X}_{m}\mathbf{P}_{m})^{\text{text}}\big)
    Q(\mathbf{U}_{c})
    Q(\mathbf{V}_{c}),
    \label{eq:activation_comp}
\end{equation}
}
where $\mathbf{Y}_{m}^{\text{text}}$ denotes the output for text activations, and the added term is the quantized compensation for the activation-induced output deviation.

\vspace{-10pt}\paragraph{Learnable Constrained Low-rank Matrix.}
Both CWS and MAC rely on low-rank branches. However, freely learned parameters, such as LoRA~\cite{hu2022lora} or QLoRA~\cite{dettmers2023qlora}, easily overfit on small 
\begin{wrapfigure}{r}{0.35\textwidth}
  \centering
  \vspace{-5pt}
  \includegraphics[width=\linewidth]{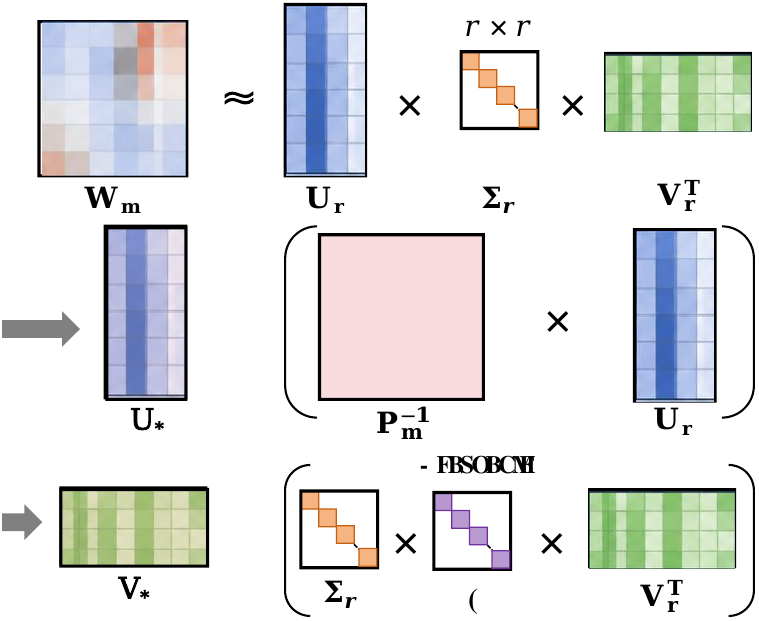}
  \vspace{-10pt}
  \caption{Construction process.}
  \label{fig:lowrank}
\end{wrapfigure}
calibration sets, while fixed SVD-based components~\cite{li2024svdquant, hu2026masquant} lack the flexibility needed to absorb quantization errors arising from modality heterogeneity. To balance these two choices, we propose a learnable constrained low-rank parameterization, as shown in Fig.~\ref{fig:lowrank}. The low-rank components in both branches are closely related to the transformed weight $\mathbf{P}_{m}^{-1}\mathbf{W}_{m}$: the smoothing branch is derived directly from it, while the compensation branch is designed to approximate it. Therefore, we build the low-rank structure on the shared transformation $\mathbf{P}_{m}^{-1}$ and the original weight $\mathbf{W}_{m}$. We first compute its rank-$r$ truncated SVD:
{\small
\begin{equation}
    \mathbf{W}_{m} \approx \mathbf{U}_{r}\mathbf{\Sigma}_{r}\mathbf{V}_{r}^{\top},
    \label{eq:truncated_svd}
\end{equation}
}
where $\mathbf{U}_{r}\in\mathbb{R}^{D_{\mathrm{in}} \times r}$ and $\mathbf{V}_{r}^{\top}\in\mathbb{R}^{r \times D_{\mathrm{out}}}$. We then construct the low-rank matrices for both branches as:
{\small
\begin{equation}
    \mathbf{U}_{*}
    =
    \mathbf{P}_{m}^{-1}\mathbf{U}_{r},
    \qquad
    \mathbf{V}_{*}
    =
    \mathbf{\Sigma}_{r}
    \mathbf{G}_{*}
    \mathbf{V}_{r}^{\top},
    \quad *\in\{s,c\}.
    \label{eq:constrained_low_rank}
\end{equation}
}

In this design, the constraint comes from anchoring the matrices to the shared transformation $\mathbf{P}_{m}^{-1}$ and the dominant SVD subspace of $\mathbf{W}_{m}$, which ensures structural consistency with the main path. Meanwhile, learnability is confined to a diagonal gating matrix $\mathbf{G}_{*}\in\mathbb{R}^{r \times r}$, allowing each branch to adaptively reweight the singular directions without introducing significant parameter overhead.

\section{Experiments}\label{sec:experi}

\subsection{Experimental Setups}
We evaluate on four VLMs from two representative architectures, including Qwen2.5-VL~\cite{bai2025qwen25vltechnicalreport} (3B/7B) and LLaVA-v1.5~\cite{liu2024improved} (7B/13B). Following recent works~\cite{li2025mbq,hu2026masquant}, we quantize only the LLM component. All methods use the same calibration samples and random seeds for fair comparison. For Qwen2.5-VL, we compare SplitQ with SmoothQuant~\cite{xiao2023smoothquant}, MBQ~\cite{li2025mbq}, and MASQuant~\cite{hu2026masquant}. For LLaVA-v1.5, we compare with Q-VLM~\cite{wang2024q}, DuQuant~\cite{lin2024duquant}, and QSVD~\cite{wang2025qsvd}. All baselines are evaluated with their recommended settings. We use MMMU~\cite{yue2023mmmu}, OCRBench~\cite{fu2024ocrbench}, TextVQA~\cite{singh2019towards}, SEED-Bench~\cite{li2023seed}, VizWiz~\cite{gurari2018vizwiz}, and ScienceQA~\cite{lu2022learn} as our benchmarks. For the transformation matrices $\mathbf{P}$ used in our method, we adopt the learnable affine form from FlatQuant~\cite{sun2024flatquant}. The smoothing and compensation branches are optional components. By default, MOCD selects 2\% modality-aware outlier channels for vision and text modalities, and the low-rank~\cite{hu2022lora} decomposition ranks of the smooth and compensation branches are set to 2\% and 3\% of the full weight rank, respectively. To preserve the representation ability of all auxiliary low-rank structures, including modality-specific independent branches as well as CWS and MAC, we apply a 4-bit quantization lower bound to these components.

\subsection{Main Results}
Table~\ref{tab:qwen25vl_combined} presents results on the Qwen2.5-VL series. At W4A8 and W4A4, SplitQ outperforms all baselines and obtains performance on par with full-precision models, while other methods suffer severe accuracy drops, particularly at W4A4. For challenging W3A3 and W3A2 settings, most existing methods cannot output normal outputs, yet SplitQ still maintains competitive performance.

Following~\cite{wang2025qsvd}, Table~\ref{tab:llava_quant_combined_rotated} further provides results on the LLaVA-v1.5 series across the same evaluation benchmarks. 
SplitQ still obtains superior results at W4A8 and W4A4. In lower-bit configurations, existing baselines suffer from unstable inference or complete failure, whereas SplitQ remains robust. These findings demonstrate the promising generalization ability of SplitQ across various VLM architectures, and verify that effectively addressing modality heterogeneity through explicit decoupling and refined calibration is important for stable low-bit multimodal quantization.

\begin{table*}[t]
\centering
\setlength{\tabcolsep}{4.2pt} 
\caption{Quantization results on Qwen2.5-VL-3B and Qwen2.5-VL-7B. Our method clearly outperforms strong baselines under all settings. `-' means no meaningful results are obtained for respective baselines.}
\resizebox{\linewidth}{!}{%
\begin{tabular}{l | c | ccccccc | ccccccc}
\toprule
& & \multicolumn{7}{c|}{Qwen2.5-VL-3B} & \multicolumn{7}{c}{Qwen2.5-VL-7B} \\
\cmidrule(lr){3-9} \cmidrule(lr){10-16}
Method & Bits & MMMU & SEED & OCRBench & VizWiz & ScienceQA & TextVQA & Avg. & MMMU & SEED & OCRBench & VizWiz & ScienceQA & TextVQA & Avg. \\
\midrule
FP16 &  W16A16
& 42.2 & 69.9 & 79.3 & 69.1 & 81.9 & 77.9 & 70.0
& 46.7 & 73.0 & 83.8 & 70.8 & 88.4 & 82.9 & 74.3 \\
\midrule
SQ & \multirow{4}{*}{W4A8} & 25.6 & 55.7 & 66.9 & 57.5 & 72.1 & 63.9 & 56.9 & 37.8 & 62.7 & 70.2 & 61.5 & 83.3 & 71.1 & 64.4 \\
MBQ & & 41.2 & 58.2 & 66.9 & 65.0 & 76.7 & 73.4 & 63.5 & 43.3 & 67.7 & 74.1 & 64.3 & 86.0 & 74.8 & 68.3 \\
MASQ & & \textbf{46.7} & 59.7 & 67.2 & 62.7 & 77.9 & 69.2 & 63.9 & 43.4 & 69.5 & 72.8 & 66.4 & 85.7 & 77.0 & 69.1 \\
\rowcolor{blue!10}SplitQ & 
& 46.3 & \textbf{69.7} & \textbf{79.1} & \textbf{67.7} & \textbf{82.4} & \textbf{77.4} & \textbf{70.4} 
& \textbf{49.1} & \textbf{73.2} & \textbf{83.5} & \textbf{68.7} & \textbf{88.1} & \textbf{82.6} & \textbf{74.2} 
\\
\midrule
SQ & \multirow{4}{*}{W4A4} 
& 23.3 & 0.0 & 0.0 & 0.0 & 0.0 & 0.0 & 3.9 
& 24.8 & 0.0 & 0.2 & 0.0 & 0.7 & 0.0 & 4.3 \\
MBQ & & 25.0 & 0.0 & 0.0 & 0.0 & 0.0 & 0.0 & 4.2 & 26.7 & 3.0 & 0.5 & 0.0 & 0.9 & 0.0 & 5.2 \\
MASQ & & 26.7 & 0.0 & 7.7 & 0.0 & 0.0 & 0.0 & 5.7 & 25.0 & 0.6 & 13.2 & 0.0 & 7.1 & 0.4 & 7.7 \\
\rowcolor{blue!10}SplitQ & 
& \textbf{43.7} & \textbf{69.3} & \textbf{78.8} & \textbf{67.6} & \textbf{81.5} & \textbf{77.0} & \textbf{69.6} 
& \textbf{46.9} & \textbf{72.6} & \textbf{83.0} & \textbf{68.4} & \textbf{87.9} & \textbf{82.5} & \textbf{73.5} \\
\midrule
SQ & \multirow{4}{*}{W3A3} 
& - & - & - & - & - & - & - 
& - & - & - & - & - & - & - \\
MBQ & & - & - & - & - & - & - & - & - & - & - & - & - & - & - \\
MASQ & & - & - & - & - & - & - & - & - & - & - & - & - & - & - \\
\rowcolor{blue!10}SplitQ & 
& \textbf{39.8} & \textbf{67.2} & \textbf{74.5} & \textbf{63.4} & \textbf{70.6} & \textbf{71.2} & \textbf{64.5} 
& \textbf{43.9} & \textbf{71.1} & \textbf{79.6} & \textbf{63.7} & \textbf{80.3} & \textbf{78.5} & \textbf{69.5} \\
\midrule
\rowcolor{blue!10}SplitQ & W3A2 
& \textbf{33.1} & \textbf{51.3} & \textbf{57.0} & \textbf{54.5} & \textbf{49.1} & \textbf{50.2} & \textbf{49.2} 
& \textbf{35.6} & \textbf{57.3} & \textbf{61.5} & \textbf{50.1} & \textbf{55.6} & \textbf{62.6} & \textbf{53.7} \\
\bottomrule
\end{tabular}%
}
\vspace{-1pt}
\label{tab:qwen25vl_combined}
\end{table*}

\begin{table*}[t]
\centering
\setlength{\tabcolsep}{3pt}
\caption{Quantization results on LLaVA-v1.5 7B and 13B.}
\resizebox{\linewidth}{!}{%
\begin{tabular}{l|c|cccc|cccc|cccc|cccc}
\toprule
\multirow{2}{*}{Model} & \multirow{2}{*}{Bits} 
& \multicolumn{4}{c|}{Duquant} 
& \multicolumn{4}{c|}{QVLM} 
& \multicolumn{4}{c|}{QSVD} 
& \multicolumn{4}{c}{SplitQ (Ours)} \\
\cline{3-18}
& & SEED-I & VizWiz & SciQA & Avg 
& SEED-I & VizWiz & SciQA & Avg 
& SEED-I & VizWiz & SciQA & Avg 
& SEED-I & VizWiz & SciQA & Avg \\
\midrule

\multirow{5}{*}{\makecell{LLaVA-v1.5\\7B}}
& FP16
& 66.2 & 54.3 & 70.0 & 63.5 
& 66.2 & 54.3 & 70.0 & 63.5 
& 66.2 & 54.3 & 70.0 & 63.5 
& 66.2 & 54.3 & 70.0 & 63.5 \\
\cmidrule(lr){2-18}
& W4A8 
& 54.4 & 50.6 & 55.3 & 53.4 
& 46.1 & 48.7 & 53.2 & 49.3 
& 57.8 & 53.5 & 63.6 & 58.3 
& \cellcolor{blue!10}65.3 & \cellcolor{blue!10}56.6 & \cellcolor{blue!10}70.3 & \cellcolor{blue!10}\textbf{64.1} \\

& W4A4 
& 51.5 & 49.8 & 54.8 & 52.0 
& 37.2 & 48.9 & 53.1 & 46.4 
& 55.1 & 53.6 & 57.7 & 55.5 
& \cellcolor{blue!10}64.8 & \cellcolor{blue!10}54.6 & \cellcolor{blue!10}69.4 & \cellcolor{blue!10}\textbf{62.9} \\

& W3A3 
& - & - & - & - 
& - & - & - & - 
& - & - & - & - 
& \cellcolor{blue!10}61.5 & \cellcolor{blue!10}57.2 & \cellcolor{blue!10}64.9 & \cellcolor{blue!10}\textbf{61.2} \\

& W3A2 
& - & - & - & - 
& - & - & - & - 
& - & - & - & - 
& \cellcolor{blue!10}54.4 & \cellcolor{blue!10}55.7 & \cellcolor{blue!10}52.3 & \cellcolor{blue!10}\textbf{54.1} \\

\midrule

\multirow{5}{*}{\makecell{LLaVA-v1.5\\13B}}
& FP16 
& 68.3 & 57.3 & 74.7 & 66.8 
& 68.3 & 57.3 & 74.7 & 66.8 
& 68.3 & 57.3 & 74.7 & 66.8 
& 68.3 & 57.3 & 74.7 & 66.8 \\
\cmidrule(lr){2-18}
& W4A8 
& 66.1 & 56.5 & 72.3 & 65.0 
& 64.2 & 55.7 & 68.4 & 62.8 
& 66.8 & 56.9 & 75.0 & 66.2 
& \cellcolor{blue!10}68.3 & \cellcolor{blue!10}57.7 & \cellcolor{blue!10}74.7 & \cellcolor{blue!10}\textbf{66.9} \\

& W4A4 
& 64.6 & 55.3 & 67.2 & 62.4 
& 48.3 & 50.9 & 65.0 & 54.7 
& 67.0 & 56.8 & 67.8 & 63.9 
& \cellcolor{blue!10}67.9 & \cellcolor{blue!10}57.2 & \cellcolor{blue!10}74.0 & \cellcolor{blue!10}\textbf{66.4} \\

& W3A3 
& - & - & - & - 
& - & - & - & - 
& - & - & - & - 
& \cellcolor{blue!10}64.6 & \cellcolor{blue!10}56.6 & \cellcolor{blue!10}69.5 & \cellcolor{blue!10}\textbf{63.6} \\

& W3A2 
& - & - & - & - 
& - & - & - & - 
& - & - & - & - 
& \cellcolor{blue!10}59.2 & \cellcolor{blue!10}57.0 & \cellcolor{blue!10}61.8 & \cellcolor{blue!10}\textbf{59.3} \\

\bottomrule
\end{tabular}%
}
\vspace{-1pt}
\label{tab:llava_quant_combined_rotated}
\end{table*}

\subsection{Ablation Study}

\vspace{-1pt}\paragraph{Channel Selection Rule.} To verify the efficacy of MOCD, we compare it with various channel selection schemes: no selection (`None'), random selection (`Random'), and heuristic methods based on the maximum or mean activations of text (`Text\_max'), vision (`Vision\_max'), or all multimodal tokens (`Multi\_max', `Multi\_mean'). All approaches select an equal number of channels for fair comparison. As shown in Table~\ref{tab:w3a3_results}, activation-driven methods consistently outperform random selection, demonstrating the importance of salient channels. More importantly, MOCD delivers the best overall results across all settings, showing that it effectively selects outlier channels. 

\vspace{-8pt}\paragraph{Cluster Number within MOCD.} We also conduct ablation experiments on the cluster number $k$ in Table~\ref{tab:w3a3_results}. The results suggest that $k=3$ yields optimal performance, which we adopt by default.

\begin{wraptable}{r}{0.45\textwidth}
    \centering
    \vspace{-10pt} 
    \caption{Similarity of MOCD Channel Selection under different calibration set size ($N$).}
    \label{tab:stability}
    \setlength{\tabcolsep}{3pt}
    \resizebox{\linewidth}{!}{
        \begin{tabular}{l|ccc}
        \toprule
        Sim. (\%)~~~~~~~~~~~~~~ & ~~~~\textbf{$32$}~~~~ & ~~~~\textbf{$64$}~~~~ & ~~~~\textbf{$256$}~~~~ \\        \midrule
        Qwen2.5-VL 7B  & 94.1 & 93.3 & 94.5 \\
        LLaVA-v1.5 7B  & 89.5 & 95.5 & 95.7 \\
        \bottomrule
        \end{tabular}
    }
    \vspace{-5pt}
\end{wraptable}
\vspace{-8pt}\paragraph{Stability of Outlier Channel Selection in MOCD.}
Practical post-training quantization (PTQ) relies on limited calibration data. To verify that MOCD remains stable under varying calibration set sizes, we measure the Jaccard similarity between outlier channels selected from calibration set of different sizes and those identified using our default calibration setting (128 samples).
As shown in Table~\ref{tab:stability}, the outlier channels selected by MOCD are stable with respect to the size of calibration set.

\begin{table*}[t]
\centering
\setlength{\tabcolsep}{6pt}
\caption{Ablation of channel selection rules and cluster number ($k$) within MOCD at W3A3. 
}
\resizebox{\linewidth}{!}{%
\begin{tabular}{l||cccc|cccc}
\toprule
\multirow{2}{*}{Method} & \multicolumn{4}{c|}{Qwen2.5-VL-3B} & \multicolumn{4}{c}{Qwen2.5-VL-7B} \\
\cmidrule{2-5} \cmidrule{6-9}
& MMMU & OCRBench & ScienceQA & Average & MMMU & OCRBench & ScienceQA & Average \\
\midrule
FP16 & 42.2 & 79.3 & 81.9 & 67.8 & 46.7 & 83.8 & 88.4 & 73.0 \\
\midrule
None & 34.0 & 66.1 & 61.6 & 53.9 & 35.1 & 73.2 & 71.5 & 59.9 \\
Random & 33.8 & 69.4 & 61.6 & 54.9 & 39.4 & 76.2 & 75.1 & 63.5 \\
Text\_max & 38.1 & 71.3 & 65.6 & 58.3 & 38.9 & 71.7 & 79.1 & 63.2 \\
Vision\_max & 37.0 & 69.8 & 68.2 & 58.3 & 39.1 & 77.3 & 77.0 & 64.4 \\
Multi\_max & 34.1 & 72.7 & 61.7 & 56.1 & 37.1 & 78.3 & 77.7 & 64.3 \\
Multi\_mean & 36.8 & 66.0 & 63.3 & 55.3 & 36.8 & 76.1 & 75.7 & 62.8 \\
\midrule
MOCD (k=1) & \textbf{37.7} & 70.2 & 68.2 & 58.4 & \textbf{40.3} & 76.3 & 77.0 & 64.5 \\
\rowcolor{blue!10}MOCD (k=3) & 37.2 & 71.1 & \textbf{69.0} & \textbf{59.1} & 40.2 & \textbf{77.8} & \textbf{78.0} & \textbf{65.3} \\
MOCD (k=5) & 37.2 & \textbf{71.5} & 68.8 & \textbf{59.1} & 40.0 & 77.2 & 77.6 & 64.9 \\
\bottomrule
\end{tabular}%
}
\vspace{-1pt}
\label{tab:w3a3_results}
\end{table*}

\begin{table*}[t]
\centering
\setlength{\tabcolsep}{5pt}
\caption{Ablation on the compensation target of MAC at W3A2.}
\resizebox{\linewidth}{!}{%
\begin{tabular}{l|c|cccc|cccc}
\toprule
\multicolumn{2}{c|}{} & \multicolumn{4}{c|}{Qwen2.5-VL-3B} & \multicolumn{4}{c}{Qwen2.5-VL-7B} \\
\cmidrule{3-6} \cmidrule{7-10}
Method & Target & MMMU & OCRBench & ScienceQA & Average & MMMU & OCRBench & ScienceQA & Average \\
\midrule
w/o MAC & / & 29.2 & 39.9 & 20.1 & 29.7 & 28.2 & 47.0 & 19.1 & 31.4 \\
\midrule
w/ MAC & vision & 31.5 & 46.6 & 28.9 & 35.7 & 29.4 & 50.0 & 42.6 & 40.6 \\
\rowcolor{blue!10}w/ MAC & text & 32.9 & 56.8 & 48.9 & 46.2 & {35.6} & \textbf{61.9} & 55.3 & \textbf{50.9} \\
w/ MAC & vision \& text & \textbf{33.0} & \textbf{57.2} & \textbf{49.6} & \textbf{46.6} & \textbf{35.6} & {61.5} & \textbf{55.6} & \textbf{50.9} \\
\bottomrule
\end{tabular}%
}
\vspace{-1pt}
\label{tab:ablation_text}
\end{table*}

\begin{table*}[!htbp]
\centering
\setlength{\tabcolsep}{5pt}
\vspace{-0pt}
\caption{Ablation study of core components in SplitQ at W3A3 and W3A2. FlatQ: FlatQuant, CWS: Cross-Modal Weight Smoothing, MAC: Modality-Specific Activation Compensation.}
\resizebox{\linewidth}{!}{%
\begin{tabular}{l|c|cccc|cccc}
\toprule
\multicolumn{2}{c|}{} & \multicolumn{4}{c|}{Qwen2.5-VL-3B} & \multicolumn{4}{c}{Qwen2.5-VL-7B} \\
\cmidrule{3-6} \cmidrule{7-10}
Method & bits & MMMU & OCRBench & ScienceQA & Average & MMMU & OCRBench & ScienceQA & Average \\
\midrule
FP16 & W16A16 & {42.2} & {79.3} & {81.9} & {67.8} & {46.7} & {83.8} & {88.4} & {73.0} \\
\midrule
Base (FlatQ) & \multirow{4}{*}{W3A3} & 34.0 & 66.1 & 61.6 & 53.9 & 35.1 & 73.2 & 71.5 & 59.9 \\
MOCD & & 37.2 & 71.1 & 69.0 & 59.1 & 40.2 & 77.8 & 78.0 & 65.3 \\
MOCD + CWS & & 37.7 & 72.7 & 69.6 & 60.0 & 41.1 & 78.2 & 78.8 & 66.0 \\
\rowcolor{blue!10}MOCD + CWS + MAC & & \textbf{39.8} & \textbf{74.5} & \textbf{70.6} & \textbf{61.7} & \textbf{43.9} & \textbf{79.6} & \textbf{80.3} & \textbf{68.0} \\
\midrule
Base (FlatQ) & \multirow{4}{*}{W3A2} & 25.5 & 0.0 & 0.0 & 8.5 & 26.0 & 2.5 & 0.04 & 9.6 \\
MOCD & & 26.6 & 21.7 & 3.5 & 17.2 & 24.9 & 40.2 & 2.4 & 22.5 \\
MOCD + CWS & & 29.2 & 39.9 & 20.1 & 29.7 & 28.2 & 47.0 & 19.1 & 31.4 \\
\rowcolor{blue!10}MOCD + CWS + MAC & & \textbf{32.9} & \textbf{56.8} & \textbf{48.9} & \textbf{46.2} & \textbf{35.6} & \textbf{61.9} & \textbf{55.3} & \textbf{50.9} \\
\bottomrule
\end{tabular}%
}
\vspace{-1pt}
\label{tab:ablation_w3a3_w3a2}
\end{table*}

\vspace{-8pt}\paragraph{Ablation on Compensation Target.}
We study which modality should be compensated in MAC by comparing four settings: without MAC, compensating only vision activations, only text activations, and both modalities. As shown in Table~\ref{tab:ablation_text}, text-only compensation yields substantially larger gains than vision-only compensation on both Qwen2.5-VL-3B and Qwen2.5-VL-7B. Moreover, compensating both modalities brings only marginal or no improvement over text-only compensation. This shows that text activations are more sensitive to quantization noise, whereas vision activations are relatively more redundant. Therefore, we adopt text-only compensation in the final design for a better trade-off between performance and inference efficiency.

\vspace{-8pt}\paragraph{Core Component Ablation Study.} Table~\ref{tab:ablation_w3a3_w3a2} ablates the core components of SplitQ at W3A3 and W3A2. ``Base'', i.e., FlatQuant, only uses the affine transformation matrix $\mathbf{P}$. 
Starting from the baseline, we gradually add our proposed techniques: MOCD, CWS, and MAC.
The results verify the effectiveness of each component. MOCD consistently improves over the baseline, indicating the necessity of modality-specific channel separation. Furthermore, the significance of the weight smoothing and activation compensation modules is highlighted under the more challenging W3A2 setting. While MOCD successfully isolates heterogeneous outliers, both CWS and MAC effectively ease the quantization difficulty of the remaining activations. These results support our design insight: while decoupling outliers is crucial, effectively smoothing the weights and compensating for activations are indispensable for preventing severe degradation at ultra-low bits.

\subsection{Inference Speedup}
Previous works~\cite{li2024svdquant,hu2026masquant} have shown that the Nunchaku\cite{li2024svdquant} efficiently supports low-rank structured
quantization. To evaluate the practical efficiency of SplitQ, we implement a customized W4A4 
\begin{wraptable}{r}{0.56\textwidth}
    \centering
    \vspace{-8pt} 
    \caption{Efficiency comparisons of different methods on Qwen2.5-VL-7B on RTX 4090 (seq. len=2048) at W4A4. MASQ: MASQuant.}
    \label{tab:speedup}
    \setlength{\tabcolsep}{3pt}
    \resizebox{\linewidth}{!}{
        \begin{tabular}{l|c|c|c|c}
        \toprule
         Method~~~ & \makecell{~~~~Prefill~~~~\\(ms)} & ~~~Speedup~~~ & \makecell{~~~Mem~~~\\(GB)} & \makecell{~~~Mem~~~\\~Saving~} \\
        \midrule
        
        FP16 & 2146.01 & - & 17.27 & - \\
        MBQ & 643.68  & 3.33$\times$ & 8.89 & 1.94$\times$ \\
        MASQ & 696.44  & 3.07$\times$ & 9.42 & 1.83$\times$ \\
        \cmidrule{1-5}
        SplitQ & 742.17 & 2.89$\times$ & 9.48 & 1.82$\times$\\
        \bottomrule
        \end{tabular}
    }
    \vspace{-5pt}
\end{wraptable}
CUDA kernel that fuses projection and quantization operators to reduce memory access. Table~\ref{tab:speedup} 
presents efficiency comparisons on Qwen2.5-VL-7B under batch size of 8. SplitQ achieves a 2.89$\times$ 
speedup over FP16 and reduces memory usage by 1.82$\times$. Compared with MBQ and MASQuant, 
SplitQ introduces only moderate latency overhead, while maintaining similar memory consumption. These results demonstrate that SplitQ remains practical for efficient VLM deployment at W4A4.

\section{Conclusion}
In this study, we identify the modality heterogeneity barrier as an important challenge for high-precision VLM quantization. To address this, we propose SplitQ, which mitigates the adverse effects of differing modal distributions through channel-level decoupling and cross-modal calibration. SplitQ achieves state-of-the-art performance across multiple benchmarks. Furthermore, our exploration into ultra-low bit scenarios demonstrates the potential for significantly higher compression ratios, providing insights for future multimodal intelligence.

Despite these promising results, SplitQ has some limitations. The introduced multi-branch computations bring additional overhead, which suggests opportunities for further hardware-aware optimizations to enhance practical efficiency.
Moreover, the modality-specific channel ratio is currently fixed, and dynamically adjusting it according to different VLM architectures and data distributions has the potential to further improve its flexibility.

\bibliographystyle{plainnat}
\bibliography{main}

\clearpage
\appendix
\section*{Appendix}

\begin{table*}[ht]
\centering
\setlength{\tabcolsep}{5pt}
\caption{Ablation study on the rank ratio of CWS.}
\resizebox{\linewidth}{!}{%
\begin{tabular}{l|c|c|cccc|cccc}
\toprule
\multicolumn{3}{c|}{} & \multicolumn{4}{c|}{Qwen2.5-VL-3B} & \multicolumn{4}{c}{Qwen2.5-VL-7B} \\
\cmidrule{4-7} \cmidrule{8-11}
Method & Rank Ratio & bits & MMMU & OCRBench & ScienceQA & Average & MMMU & OCRBench & ScienceQA & Average \\
\midrule
FP16 & - & W16A16 & {42.2} & {79.3} & {81.9} & {67.8} & {46.7} & {83.8} & {88.4} & {73.0} \\
\midrule
 & 0.01 & & 38.9 & 73.3 & 69.9 & 60.7 & 42.1 & 77.1 & 78.6 & 65.9 \\
SplitQ & 0.02 & W3A3 & 39.8 & 74.5 & 70.6 & 61.7 & 43.9 & 79.6 & 80.3 & 68.0 \\
 & 0.03 & & 39.6 & 74.6 & 71.1 & 61.7 & 43.2 & 78.4 & 80.5 & 67.3\\
\midrule
& 0.01 & & 32.4 & 53.5 & 48.2 & 44.7 & 33.6 & 57.3 & 53.3 & 48.0\\
SplitQ & 0.02 & W3A2 & 33.1 & 57.0 & 49.1 & 46.4 & 35.6 & 61.5 & 55.6 & 50.9 \\
& 0.03 & & 34.0 & 57.1 & 50.5 & 47.2 & 36.0 & 61.1 & 55.8 & 51.0\\
\bottomrule
\end{tabular}%
}
\label{tab:ablation_CWSrank}
\end{table*}

\begin{table*}[ht]
\centering
\setlength{\tabcolsep}{5pt}
\caption{Ablation study on the rank ratio of MAC.}
\resizebox{\linewidth}{!}{%
\begin{tabular}{l|c|c|cccc|cccc}
\toprule
\multicolumn{3}{c|}{} & \multicolumn{4}{c|}{Qwen2.5-VL-3B} & \multicolumn{4}{c}{Qwen2.5-VL-7B} \\
\cmidrule{4-7} \cmidrule{8-11}
Method & Rank Ratio & bits & MMMU & OCRBench & ScienceQA & Average & MMMU & OCRBench & ScienceQA & Average \\
\midrule
FP16 & - & W16A16 & {42.2} & {79.3} & {81.9} & {67.8} & {46.7} & {83.8} & {88.4} & {73.0} \\
\midrule
 & 0.01 & & 38.4 & 73.6 & 69.9 & 60.6 & 42.1 & 78.5 & 79.2 & 66.4 \\
SplitQ & 0.03 & W3A3 & 39.8 & 74.5 & 70.6 & 61.7 & 43.9 & 79.6 & 80.3 & 68.0 \\
 & 0.05 & & 38.9 & 73.9 & 70.8 & 61.2 & 42.0 & 78.2 & 79.8 & 66.7\\
\midrule
& 0.01 & & 31.2 & 49.7 & 35.5 & 38.8 & 33.1 & 55.8 & 40.9 & 43.3 \\
SplitQ & 0.03 & W3A2 & 33.1 & 57.0 & 49.1 & 46.4 & 35.6 & 61.5 & 55.6 & 50.9 \\
& 0.05 & & 34.4 & 58.9 & 51.2 & 48.1 & 36.8 & 63.7 & 57.0 & 52.5\\
\bottomrule
\end{tabular}%
}
\label{tab:ablation_MACrank}
\end{table*}

\begin{table*}[ht]
\centering
\setlength{\tabcolsep}{5pt}
\caption{Ablation of fixed and learnable low-rank matrix designs for the CWS module at W3A3 on Qwen2.5-VL.}
\resizebox{\linewidth}{!}{%
\begin{tabular}{l|c|c|cccc|cccc}
\toprule
\multicolumn{3}{c|}{} & \multicolumn{4}{c|}{Qwen2.5-VL-3B} & \multicolumn{4}{c}{Qwen2.5-VL-7B} \\
\cmidrule{4-7} \cmidrule{8-11}
Method & setting & bits & MMMU & OCRBench & ScienceQA & Average & MMMU & OCRBench & ScienceQA & Average \\
\midrule
FP16 & - & W16A16 & {42.2} & {79.3} & {81.9} & {67.8} & {46.7} & {83.8} & {88.4} & {73.0} \\
\midrule
& w/o CWS & & 37.5 & 72.2 & 69.4 & 59.7 & 40.5 & 78.8 & 78.2 & 65.8 \\
SplitQ & fixed & W3A3 & 35.3 & 69.1 & 65.8 & 56.7 & 40.1 & 74.6 & 78.3 & 64.3 \\
& learnable & & \textbf{39.8} & \textbf{74.5} & \textbf{70.6} & \textbf{61.7} & \textbf{43.9} & \textbf{79.6} & \textbf{80.3} & \textbf{68.0}\\
\bottomrule
\end{tabular}%
}
\label{tab:cws_learnable}
\end{table*}

\begin{table*}[ht]
\centering
\setlength{\tabcolsep}{5pt}
\caption{Ablation of fixed and learnable low-rank matrix designs for the MAC module at W3A3 on Qwen2.5-VL.}
\resizebox{\linewidth}{!}{%
\begin{tabular}{l|c|c|cccc|cccc}
\toprule
\multicolumn{3}{c|}{} & \multicolumn{4}{c|}{Qwen2.5-VL-3B} & \multicolumn{4}{c}{Qwen2.5-VL-7B} \\
\cmidrule{4-7} \cmidrule{8-11}
Method & setting & bits & MMMU & OCRBench & ScienceQA & Average & MMMU & OCRBench & ScienceQA & Average \\
\midrule
FP16 & - & W16A16 & {42.2} & {79.3} & {81.9} & {67.8} & {46.7} & {83.8} & {88.4} & {73.0} \\
\midrule
& w/o MAC & & 37.7 & 72.7 & 69.6 & 60.0 & 41.1 & 78.2 & 78.8 & 66.0 \\
SplitQ & fixed & W3A3 & 36.9 & 72.0 & 67.9 & 58.9 & 40.5 & 76.4 & 77.5 & 64.4 \\
& learnable & & \textbf{39.8} & \textbf{74.5} & \textbf{70.6} & \textbf{61.7} & \textbf{43.9} & \textbf{79.6} & \textbf{80.3} & \textbf{68.0}\\
\bottomrule
\end{tabular}%
}
\label{tab:mac_learnable}
\end{table*}

\begin{table}[t]
    \centering
    \caption{End-to-end prefill-stage performance of Qwen2.5-VL-7B on RTX 4090 (seq len=2048, W4A4). Full: MOCD+CWS+MAC, BS: Batch Size.}
    \label{tab:speedup_full}
    \setlength{\tabcolsep}{3pt}
    \resizebox{0.6\linewidth}{!}{%
        \begin{tabular}{c|l|c|c|c|c|c}
        \toprule
        BS & Method & setting & \makecell{Prefill\\(ms)} & Speedup & \makecell{Mem\\(GB)} & \makecell{Mem\\Saving} \\
        \midrule
        \multirow{6}{*}{1}
        & FP16 & - & 191.82 & / & 13.73 & / \\
        \cmidrule{2-7}
        & SplitQ & MOCD & 70.99 & 2.70$\times$ & 4.85 & 2.83$\times$ \\
        & SplitQ & MOCD+CWS & 75.55 & 2.53$\times$ & 5.05 & 2.71$\times$\\
        & SplitQ & Full & 81.78 & 2.34$\times$ & 5.42 & 2.53$\times$\\
        \midrule
        \multirow{6}{*}{8}
        & FP16 & - & 2146.01 & / & 17.27 & / \\
        \cmidrule{2-7}
        & SplitQ & MOCD & 648.13 & 3.31$\times$ & 8.89 &  1.94$\times$ \\
        & SplitQ & MOCD+CWS & 693.58 & 3.09$\times$ & 9.18 & 1.88$\times$ \\
        & SplitQ & Full & 742.17 & 2.89$\times$ & 9.48 & 1.82$\times$\\
        \bottomrule
        \end{tabular}
    }
\end{table}

\section{Technical Appendices and Supplementary Material}

\subsection{Ablation on the Rank of CWS and MAC}
As shown in Table~\ref{tab:ablation_CWSrank} and~\ref{tab:ablation_MACrank}, we evaluate the CWS and MAC modules under different rank ratios in both W3A3 and W3A2 settings. For CWS, we test rank ratios of 0.01, 0.02, and 0.03. At W3A3, the best results are obtained with a rank ratio of 0.02 on both Qwen2.5-VL-3B and Qwen2.5-VL-7B, while at W3A2, a larger ratio of 0.03 yields better performance. For MAC, we evaluate rank ratios of 0.01, 0.03, and 0.05. The results show that 0.03 performs best at W3A3, and 0.05 achieves the best performance at W3A2. Overall, these results suggest that a moderate rank ratio is sufficient for CWS, while MAC benefits from a slightly larger rank ratio in the more aggressive quantization setting.

\subsection{Learnable Low-Rank Matrix in CWS and MAC}
To validate the rationality of our constrained learnable low-rank design in CWS and MAC, we conduct ablation experiments on different low-rank parameterization schemes. As shown in Table \ref{tab:cws_learnable} and Table \ref{tab:mac_learnable}, the traditional fixed SVD-based low-rank design achieves even worse performance than the baseline without CWS/MAC modules. This phenomenon is consistent with the motivation of our cross-modal calibration module. Fixed SVD decomposition is widely applied in single-modal compression tasks, including pure text LLMs and visual diffusion models, which is well suited to the unified data distribution of single modalities. However, vision-text multimodal models suffer from inherent heterogeneous distribution gaps. The static fixed low-rank structure lacks adaptive flexibility for cross-modal quantization noise, which instead introduces extra optimization bias and degrades quantization performance. In contrast, our learnable constrained low-rank design retains the structural prior of SVD while enabling adaptive reweighting of modal-sensitive features. It achieves consistent performance gains over both fixed SVD and module-free baselines.

\subsection{Inference Efficiency Ablation}
We further ablate the inference efficiency overhead of each component in SplitQ. Table~\ref{tab:speedup_full} summarizes the prefill latency, speedup and memory consumption. Gradually incorporating additional modules slightly increases latency and memory usage, yet all configurations achieve substantial acceleration and memory savings compared with the FP16 baseline. The results confirm that our modules introduce marginal overhead, and the full SplitQ remains efficient for practical VLM deployment.

\begin{table}[tb]
\centering
\caption{Visual encoder and LLM quantization performance on MMMU, VizWiz, and ScienceQA on Qwen2.5-VL-7B.}
\label{tab:vit_vlm_ablation}
\begin{tabular}{cccccc}
\toprule
Visual Encoder & LLM & MMMU ($\uparrow$) & VizWiz ($\uparrow$) & ScienceQA ($\uparrow$) & Average ($\uparrow$)\\
\midrule
FP16 & FP16 & 46.7 & 70.8 & 88.4 & 68.6 \\
\midrule
FP16 & W4A4 & 46.9 & 68.4 & 87.9 & 67.7 \\
FP16 & W3A3 & 43.9 & 63.7 & 80.3 & 62.6 \\
FP16 & W3A2 & 35.6 & 50.1 & 55.6 & 47.1 \\
\midrule
W4A4 & W4A4 & 46.6 & 68.0 & 88.0 & 67.5 \\
W4A4 & W3A3 & 43.3 & 63.9 & 80.3 & 62.5 \\
W4A4 & W3A2 & 35.2 & 49.7 & 53.9 & 46.3 \\
\bottomrule
\end{tabular}
\end{table}

\subsection{Visual Encoder Quantization}
SplitQ can also be extended to the visual encoder. As shown in Table~\ref{tab:vit_vlm_ablation}, we evaluate different quantization configurations for both the visual encoder and the LLM on Qwen2.5-VL-7B. The results show that quantizing the visual encoder with W4A4 preserves performance well, and the overall accuracy remains comparable to the corresponding settings with an FP16 visual encoder. This demonstrates that SplitQ is also effective for visual encoder quantization.

\subsection{Case Study for SplitQ}
For the case study, we randomly select examples from the ScienceQA test set to demonstrate the superior performance of our SplitQ over MBQ and MASQuant under quantization.

\begin{itemize}
    \item In Case 3, MBQ and MASQuant yield flawed reasoning and incorrect answers at W4A8, while SplitQ maintains accurate predictions across different bit-width configurations.
    \item At ultra-low bit settings such as W3A2, SplitQ produces stable and reliable outputs in Case 2, Case 3, Case 6, and Case 7.
    \item In the challenging Case 4, the FP16 baseline draws an incorrect conclusion, whereas SplitQ at W4A8 obtains the correct answer with reasonable reasoning. By contrast, all compared methods fail to reason correctly in Case 8.
\end{itemize}

\begin{tcolorbox}[casebox, title={\bfseries Case 1: Qwen2.5-VL 7B}]
\raggedright

\textbf{Question:} Think about the magnetic force between the magnets in each pair. Which of the following statements is true?

\begin{enumerate}[label=\Alph*., noitemsep, align=left]
    \item The magnitude of the magnetic force is the same in both pairs.
    \item The magnitude of the magnetic force is smaller in Pair 1.
    \item The magnitude of the magnetic force is smaller in Pair 2. \textcolor{gtred}{\textbf{(GT Answer)}}
\end{enumerate}

\vspace{5pt}

{
\centering
\includegraphics[width=0.7\textwidth]{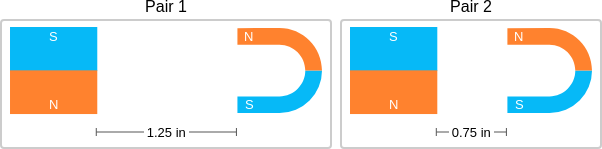}\par
}

\vspace{5pt}
\raggedright

\hrule height 0.3pt
\vspace{3pt}
\noindent\textbf{FP16 Response:} 1. Compare the magnetic force in each pair. Pair 1 has a larger distance (1.25 in) than Pair 2 (0.75 in). Magnetic force decreases with distance, so the magnitude is smaller in Pair 1.

\vspace{3pt}
\hrule height 0.3pt
\vspace{3pt}
\noindent\textbf{MBQ W4A8 Response:} The distance in Pair 1 is larger, so the magnetic force is smaller.

\noindent\textbf{MASQuant W4A8 Response:} To solve this, I need to consider both magnet size and distance. Pair 1 has a wider gap but the U-shaped magnet is larger. Pair 2 has a smaller gap, but the U-magnet is noticeably smaller. Magnet size dominates here: the smaller magnet in Pair 2 leads to weaker force overall, even with the closer distance. So \textcolor{gtred}{the magnitude is smaller in Pair 2}.

\vspace{3pt}
\hrule height 0.3pt
\vspace{3pt}
\noindent\textbf{SplitQ W4A8 Response:} Key details from the diagram: both pairs share the same rectangular magnet, but Pair 2's U-shaped magnet is smaller. Pair 1's gap is wider, but the U-magnet is bigger. \textcolor{gtred}{Smaller magnets exert weaker force, so Pair 2 has smaller force}.

\noindent\textbf{SplitQ W4A4 Response:} Magnetic force depends on magnet size and distance. \textcolor{gtred}{Pair 2's U-magnet is smaller, so even though it's closer, the force is smaller}.

\noindent\textbf{SplitQ W3A3 Response:} The magnets in Pair 1 are farther apart (1.25 in vs 0.75 in). Magnetic force decreases with distance, so Pair 1 has smaller force.

\vspace{5pt}
\hrule height 0.3pt
\vspace{5pt}
\noindent\textbf{Note:} MBQ and MASQuant W4A4 output nothing or random numbers, failing to generate valid responses.
\end{tcolorbox}

\begin{tcolorbox}[casebox, title={\bfseries Case 2: Qwen2.5-VL 7B}]
\raggedright

\textbf{Question:} What is the capital of Mississippi?

\begin{enumerate}[label=\Alph*., noitemsep, align=left]
    \item Detroit
    \item Jackson \textcolor{gtred}{\textbf{(GT Answer)}}
    \item Biloxi
    \item Topeka
\end{enumerate}

\vspace{5pt}

{
\centering
\includegraphics[width=0.5\textwidth]{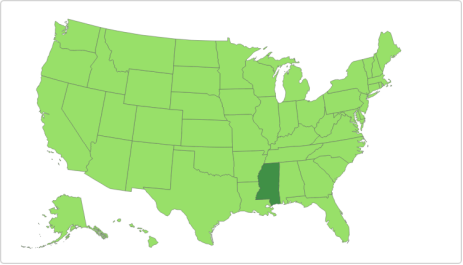}\par
}

\vspace{5pt}
\raggedright

\hrule height 0.3pt
\vspace{3pt}
\noindent\textbf{FP16 Response:} Looking at the U.S. map where Mississippi is highlighted, the capital of Mississippi is \textcolor{gtred}{Jackson}.

\vspace{3pt}
\hrule height 0.3pt
\vspace{3pt}
\noindent\textbf{MBQ W4A8 Response:} The capital of Mississippi is \textcolor{gtred}{Jackson}.

\noindent\textbf{MASQuant W4A8 Response:} Referencing the provided map and U.S. state capital knowledge, \textcolor{gtred}{Jackson is the confirmed capital of Mississippi}. Detroit is in Michigan, Biloxi is a city in Mississippi but not the capital, and Topeka is in Kansas.

\vspace{3pt}
\hrule height 0.3pt
\vspace{3pt}
\noindent\textbf{SplitQ W4A8 Response:} The highlighted state is Mississippi, and \textcolor{gtred}{its capital is Jackson}.

\noindent\textbf{SplitQ W4A4 Response:} \textcolor{gtred}{Jackson} is the capital of Mississippi.

\noindent\textbf{SplitQ W3A3 Response:} \textcolor{gtred}{B. Jackson}.

\end{tcolorbox}

\begin{tcolorbox}[casebox, title={\bfseries Case 3: Qwen2.5-VL 7B}]
\raggedright

\textbf{Question:} Which of these organisms contains matter that was once part of the lichen?

\begin{enumerate}[label=\Alph*., noitemsep, align=left]
    \item mushroom \textcolor{gtred}{\textbf{(GT Answer)}}
    \item short-tailed weasel
    \item brown lemming
    \item rough-legged hawk
    \item bilberry
\end{enumerate}

\vspace{5pt}

{
\centering
\includegraphics[width=0.5\textwidth]{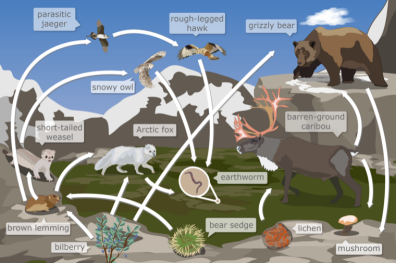}\par
}

\vspace{5pt}
\raggedright

\hrule height 0.3pt
\vspace{3pt}
\noindent\textbf{FP16 Response:} Looking at the arrows in the food web, matter from the lichen moves to the barren-ground caribou first, then can reach the mushroom either directly or via the grizzly bear. No other organisms like the bilberry, brown lemming, short-tailed weasel, or rough-legged hawk have valid paths from the lichen. \textcolor{gtred}{So the answer is mushroom}.

\vspace{3pt}
\hrule height 0.3pt
\vspace{3pt}
\noindent\textbf{MBQ W4A8 Response:} The short-tailed weasel is connected to the brown lemming, which seems to be linked to the lichen in the diagram. Thus, it should contain matter from the lichen.

\noindent\textbf{MASQuant W4A8 Response:} The brown lemming feeds on plants including the lichen, so it would take in matter from the lichen directly.

\vspace{3pt}
\hrule height 0.3pt
\vspace{3pt}
\noindent\textbf{SplitQ W4A8 Response:} \textcolor{gtred}{The mushroom is the only organism} connected to the lichen through the barren-ground caribou in the food web. It is the only one that contains matter from the lichen.

\noindent\textbf{SplitQ W4A4 Response:} \textcolor{gtred}{The mushroom has a direct path from the lichen via the caribou}, so it contains the matter.

\noindent\textbf{SplitQ W3A3 Response:} The correct organism that contains matter from \textcolor{gtred}{the lichen is the mushroom}.

\end{tcolorbox}

\begin{tcolorbox}[casebox, title={\bfseries Case 4: Qwen2.5-VL 7B}]
\raggedright

\textbf{Question:} Compare the average kinetic energies of the particles in each sample. Which sample has the higher temperature?

\begin{enumerate}[label=\Alph*., noitemsep, align=left]
    \item sample A
    \item sample B \textcolor{gtred}{\textbf{(GT Answer)}}
    \item neither; the samples have the same temperature
\end{enumerate}

\vspace{5pt}

{
\centering
\includegraphics[width=0.45\textwidth]{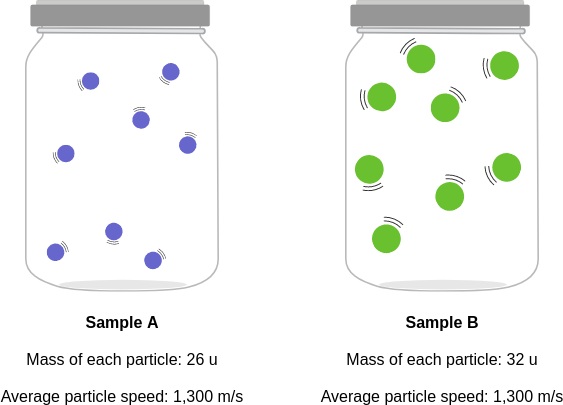}\par
}

\vspace{5pt}
\raggedright

\hrule height 0.3pt
\vspace{3pt}
\noindent\textbf{FP16 Response:} To compare the average kinetic energies and temperatures of the two samples, we analyze the particle properties. The average particle speed is the same in both Sample A and Sample B. Since temperature is determined by average kinetic energy, and kinetic energy is closely related to particle speed, the two samples should have the same average kinetic energy. Therefore, their temperatures are equal.

\vspace{3pt}
\hrule height 0.3pt
\vspace{3pt}
\noindent\textbf{MBQ W4A8 Response:} Same speed temperature equal mass higher sample B kinetic...

\noindent\textbf{MASQuant W4A8 Response:} Looking at the samples, they have the same particle speed, so their temperatures are equal.

\vspace{3pt}
\hrule height 0.3pt
\vspace{3pt}
\noindent\textbf{SplitQ W4A8 Response:} \textcolor{gtred}{Sample B has the same average speed as Sample A but a larger particle mass, so it has higher average kinetic energy and higher temperature}.

\noindent\textbf{SplitQ W4A4 Response:} Since the average particle speed is the same in both samples, their average kinetic energies are equal, so the temperatures are the same.

\noindent\textbf{SplitQ W3A3 Response:} The particle speeds are equal, so the temperatures are the same.

\vspace{5pt}
\hrule height 0.3pt
\vspace{5pt}
\noindent\textbf{Note:} MBQ and MASQuant W4A4 output nothing or random numbers, failing to generate valid responses.
\end{tcolorbox}

\begin{tcolorbox}[casebox, title={\bfseries Case 5: Qwen2.5-VL 3B}]
\raggedright

\textbf{Question:} Which of the following best describes a population in a forest in the Cascade mountains?

\begin{enumerate}[label=\Alph*., noitemsep, align=left]
    \item the western red cedar trees \textcolor{gtred}{\textbf{(GT Answer)}}
    \item the great horned owls and the northern pygmy owls
    \item the Douglas fir trees and the long-eared owls
\end{enumerate}

\vspace{5pt}

{
\centering
\includegraphics[width=0.45\textwidth]{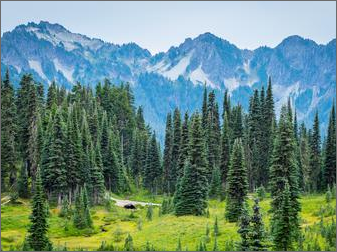}\par
}

\vspace{5pt}
\raggedright

\hrule height 0.3pt
\vspace{3pt}
\noindent\textbf{FP16 Response:} A population refers to all individuals of the same species living in a specific area. In this case, \textcolor{gtred}{the western red cedar trees are a single species in the forest}, so they form a population. The other options include multiple different species, which describe communities rather than populations. 

\vspace{3pt}
\hrule height 0.3pt
\vspace{3pt}
\noindent\textbf{MBQ W4A8 Response:} We can judge the population range by animals living in the same mountain forest area. The great horned owls and northern pygmy owls both inhabit the Cascade mountains forest, so they belong to the same group and form a population there.

\noindent\textbf{MASQuant W4A8 Response:} A population consists of one species only. \textcolor{gtred}{The western red cedar trees are a single species}, making them a population. The other options include multiple species, so they are not populations.

\vspace{3pt}
\hrule height 0.3pt
\vspace{3pt}
\noindent\textbf{SplitQ W4A8 Response:} \textcolor{gtred}{The western red cedar trees are all the same species living in the forest}, so they meet the definition of a population.

\noindent\textbf{SplitQ W4A4 Response:} The correct description of a population is \textcolor{gtred}{the western red cedar trees, as they are a single species}.

\noindent\textbf{SplitQ W3A3 Response:} The two types of owls are both birds living in the same forest, so they form a population together.

\end{tcolorbox}

\begin{tcolorbox}[casebox, title={\bfseries Case 6: Qwen2.5-VL 3B}]
\raggedright

\textbf{Question:} Which better describes the Jardines de la Reina National Park ecosystem?

\begin{enumerate}[label=\Alph*., noitemsep, align=left]
    \item It has bright sunlight. It also has shallow water. \textcolor{gtred}{\textbf{(GT Answer)}}
    \item It has water with not much salt. It also has only a few types of organisms.
\end{enumerate}

\vspace{5pt}

{
\centering
\includegraphics[width=0.45\textwidth]{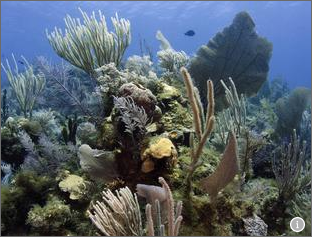}\par
}

\vspace{5pt}
\raggedright

\hrule height 0.3pt
\vspace{3pt}
\noindent\textbf{FP16 Response:} Looking at the coral reef ecosystem in the image, tropical coral reefs like this are characterized by \textcolor{gtred}{shallow, clear water with bright sunlight, which supports diverse coral and marine life}. Option A matches these key features, while Option B incorrectly describes low salinity and limited biodiversity—both of which are not true of coral reef ecosystems.

\vspace{3pt}
\hrule height 0.3pt
\vspace{3pt}
\noindent\textbf{MBQ W4A8 Response:} The ecosystem shown is a coral reef, which has \textcolor{gtred}{shallow water and bright sunlight}. The correct description is Option A.

\noindent\textbf{MASQuant W4A8 Response:} The image appears to show darker water with relatively few visible organisms, so the ecosystem may have limited sunlight and low biodiversity.

\vspace{3pt}
\hrule height 0.3pt
\vspace{3pt}
\noindent\textbf{SplitQ W4A8 Response:} The Jardines de la Reina National Park ecosystem shown is a coral reef, \textcolor{gtred}{which has shallow water and bright sunlight as described in Option A}.

\noindent\textbf{SplitQ W4A4 Response:} To determine the correct habitat of the coral reef ecosystem, we need to analyze the environmental conditions that corals depend on for survival. Corals maintain a symbiotic relationship with photosynthetic algae, which require sufficient sunlight to produce nutrients. Deep ocean areas cannot provide enough light, while shallow coastal waters allow sunlight to penetrate easily. The coral reef ecosystem has \textcolor{gtred}{shallow water and bright sunlight}, which fully satisfies the living and growth conditions of corals and their symbiotic algae, so the answer is A.

\noindent\textbf{SplitQ W3A3 Response:} Based on the coral reef in the image, \textcolor{gtred}{the ecosystem has bright sunlight and shallow water}.

\end{tcolorbox}

\begin{tcolorbox}[casebox, title={\bfseries Case 7: Qwen2.5-VL 3B}]
\raggedright

\textbf{Question:} What is the name of the colony shown?

\begin{enumerate}[label=\Alph*., noitemsep, align=left]
    \item Massachusetts
    \item Maryland
    \item North Carolina
    \item New Hampshire \textcolor{gtred}{\textbf{(GT Answer)}}
\end{enumerate}

\vspace{5pt}

{
\centering
\includegraphics[width=0.3\textwidth]{figs/q7.png}\par
}

\vspace{5pt}
\raggedright

\hrule height 0.3pt
\vspace{3pt}
\noindent\textbf{FP16 Response:} \textcolor{gtred}{New Hampshire}.

\vspace{3pt}
\hrule height 0.3pt
\vspace{3pt}
\noindent\textbf{MBQ W4A8 Response:} The colony shown in the map is in the northeast, \textcolor{gtred}{which corresponds to New Hampshire}.

\noindent\textbf{MASQuant W4A8 Response:} The map highlights a colony in the New England region. Based on the position relative to other colonies, \textcolor{gtred}{this is New Hampshire}, not Massachusetts or the southern colonies listed.

\vspace{3pt}
\hrule height 0.3pt
\vspace{3pt}
\noindent\textbf{SplitQ W4A8 Response:} The northeastern colony shown is \textcolor{gtred}{New Hampshire}.

\noindent\textbf{SplitQ W4A4 Response:} \textcolor{gtred}{New Hampshire} is the name of the colony shown in the map.

\noindent\textbf{SplitQ W3A3 Response:} The highlighted colony in the northeast is \textcolor{gtred}{New Hampshire}.

\end{tcolorbox}

\begin{tcolorbox}[casebox, title={\bfseries Case 8: Qwen2.5-VL 3B}]
\raggedright

\textbf{Question:} What can Shivani and Austin trade to each get what they want?

\begin{enumerate}[label=\Alph*., noitemsep, align=left]
    \item Austin can trade his almonds for Shivani's tomatoes.
    \item Shivani can trade her tomatoes for Austin's broccoli. \textcolor{gtred}{\textbf{(GT Answer)}}
    \item Austin can trade his broccoli for Shivani's oranges.
    \item Shivani can trade her tomatoes for Austin's carrots.
\end{enumerate}

\vspace{5pt}

{
\centering
\includegraphics[width=0.7\textwidth]{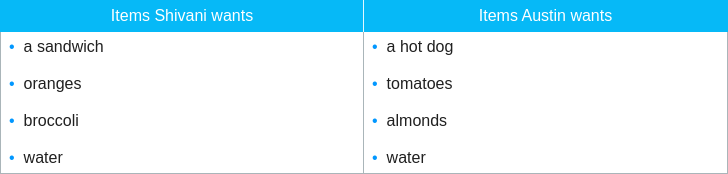}\par
}

\vspace{5pt}
\raggedright

\hrule height 0.3pt
\vspace{3pt}
\noindent\textbf{FP16 Response:} Looking at the items each person wants, Shivani needs broccoli and Austin needs tomatoes. Austin has almonds available, so trading them for Shivani's tomatoes will let him get what he wants, making this a valid trade for both.

\vspace{3pt}
\hrule height 0.3pt
\vspace{3pt}
\noindent\textbf{MBQ W4A8 Response:} Austin wants tomatoes, so he can trade his almonds for Shivani's tomatoes to get what he needs.

\noindent\textbf{MASQuant W4A8 Response:} By checking what each person wants and what items they own in the given table, we can figure out a reasonable trade. Austin hopes to get tomatoes, while Shivani needs other daily food items. Based on the table, the only way both get their desired items is if Austin trades his almonds for Shivani's tomatoes.

\vspace{3pt}
\hrule height 0.3pt
\vspace{3pt}
\noindent\textbf{SplitQ W4A8 Response:} Austin wants tomatoes, and Shivani can provide them. Since Austin has almonds, he can trade his almonds for Shivani’s tomatoes, which would let Austin get what he wants. Therefore, the answer is A.

\noindent\textbf{SplitQ W4A4 Response:} Austin can trade his almonds for Shivani's tomatoes, which works for both.

\noindent\textbf{SplitQ W3A3 Response:} The trade where Austin gives almonds to get tomatoes from Shivani is the right choice.

\end{tcolorbox}



\end{document}